\documentclass{article}
\usepackage{ijcai26}

\usepackage{times}
\usepackage{placeins}

\usepackage{soul}
\usepackage{url}
\usepackage[hidelinks]{hyperref}
\usepackage[utf8]{inputenc}
\usepackage[small]{caption}
\usepackage{graphicx}
\usepackage{amsmath}
\usepackage{amsthm}
\usepackage{booktabs}
\usepackage{algorithm}
\usepackage{algorithmic}
\usepackage[switch]{lineno}
\usepackage{etoolbox}
\usepackage{fvextra}
\usepackage[most]{tcolorbox}
\usepackage{amsmath}
\usepackage{amsfonts,amssymb}
\usepackage{mathrsfs}
\usepackage{multirow}
\usepackage{makecell}

\usepackage{pgfplots}
\usepackage{newfloat}
\usepackage{listings}
\usepackage{subfigure}
\usepackage{enumitem}
\pgfplotsset{compat=1.18}
\usepackage{wrapfig}
\usepgfplotslibrary{groupplots}
\pgfplotsset{compat=1.17} 
\usepackage{mathtools}
\definecolor{darkteal}{RGB}{0, 128, 128} 
\definecolor{lightteal}{RGB}{128, 220, 220} 

\linenumbers

\title{\textsc{StressEval}: Failure-Driven Dynamic Benchmarking for Knowledge-Intensive Reasoning in Large Language Models}

\author{
Yongrui Chen$^1$\and
Yangyang Ma$^1$\and
Xiaoying Huang$^{4}$\and
Shenyu Zhang$^1$\and\\
Huajun Chen$^2$\and
Haofen Wang$^3$\And
Guilin Qi$^{1,}$\thanks{Corresponding author}
\affiliations
$^1$Southeast University \quad $^2$Zhejiang University\\
$^3$Tongji University \quad $^4$Hangzhou Dianzi University\\
\emails
\{yongruichen, gqi\}@seu.edu.cn.com
}

\begin{document}

\maketitle

\begin{abstract}
Static benchmarks for LLMs are increasingly compromised by contamination and overfitting, especially on knowledge-intensive reasoning tasks. While recent dynamic benchmarks can alleviate staleness, they often increase difficulty at the expense of answerability and controllability.
In this paper, we propose \textsc{StressEval}, a failure-driven data synthesis framework that turns observed model failures into dynamic, challenging, and controlable test instances. \textsc{StressEval} consists of three stages: (i) it constructs a semi-structured difficulty card that identifies the failed reasoning step and its root cause; (ii) it applies a dual-perspective instance-synthesis method that targets both knowledge gaps and reasoning breakdowns while preserving the underlying difficulty factors; and (iii) it applies a gating mechanism to retain only grounded, unambiguous instances. 
Seeding from multiple knowledge-intensive reasoning datasets, we employ \textsc{StressEval} to build \textsc{Dynamic-OneEval}\footnote{\url{http://oneeval.openkg.cn}}, a focused suite of challenging dynamic benchmark. Across several state-of-the-art LLMs, \textsc{Dynamic-OneEval} yields substantially larger performance drops than the original benchmarks while retaining explicit difficulty factors, enabling more actionable iteration.
\end{abstract}

\section{Introduction}

Large language models (LLMs) have advanced rapidly, delivering strong performance across a broad range of language understanding tasks and increasingly sophisticated forms of reasoning~\cite{brown2020language,openai2023gpt4}. However, as model architectures and training pipelines evolve at pace, conventional static evaluation paradigms~\cite{chen2021evaluating,DBLP:conf/iclr/HendrycksBBZMSS21,yue2024mmmu} are becoming less informative. Leaderboard gains can obscure brittle behaviors, while widespread benchmark reuse and dataset contamination make it difficult to disentangle genuine generalization from memorization or overfitting~\cite{lin2024rethinking,wang2025rankings,kandpal2024mind,DBLP:journals/tmlr/AnwarSRPTHLJCSE24}. These limitations are especially pronounced for knowledge-intensive reasoning like \textsc{OneEval}~\cite{chen2025oneeval}, where success hinges on up-to-date factual knowledge, careful grounding, and reliable multi-step inference~\cite{jiang2024evaluating,zhou2025credible}.

\begin{figure}
\centering
	\includegraphics[width=0.48\textwidth]{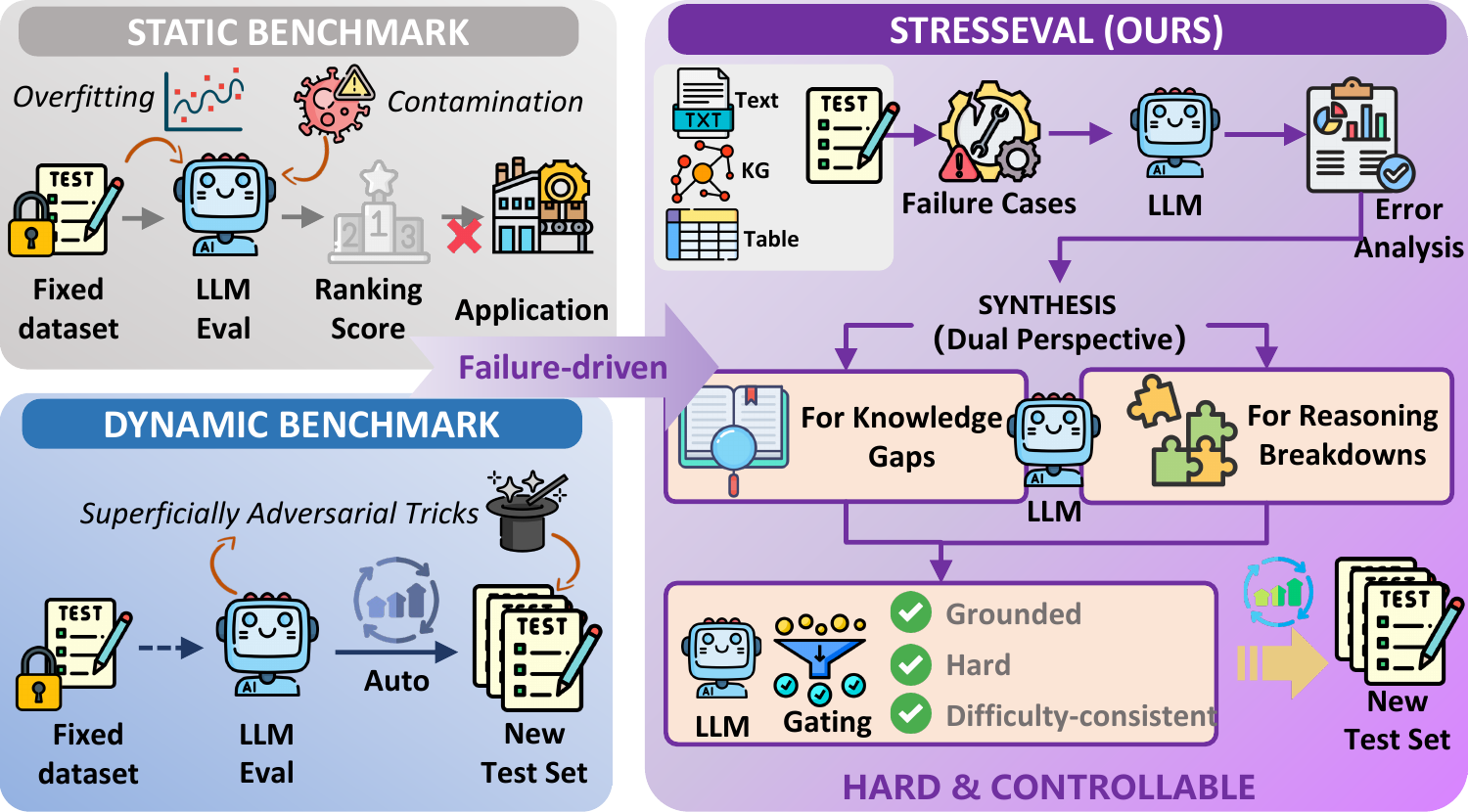}
	\caption{Upper left: static benchmarks degrade via overfitting and contamination. Bottom left: dynamic benchmarks refresh via generation but drift toward superficial trickiness. Right: \textsc{StressEval} turns observed failures into hard and controllable test instances.} \label{fig:example}
\end{figure}

Recent work~\cite{white2024livebench,zhao2025auto} has begun to explore \textit{dynamic benchmarks} that leverage automatic data synthesis to keep evaluations continually refreshed and more resistant to benchmark contamination~\cite{white2024livebench,li2024forewarned}. 
While this direction can mitigate staleness, automatically generated instances often drift toward ungrounded content or superficially adversarial tricks, rather than isolating a well-defined capability or failure mode~\cite{white2024livebench,li2024forewarned,wang2025rankings} (see the bottom-left of Figure~\ref{fig:example}). Consequently, the revealed errors are difficult to interpret and even harder to translate into actionable mitigations for model designers. 
This raises an open problem: how to generate evaluation instances that are both (i) \textit{challenging}: they reliably stress specific knowledge points or reasoning patterns where models are weak, and (ii) \textit{controllable difficulty}: their construction is governed by explicit factors so that failures can be traced back to concrete causes and directly guide model or system improvements.


To address this gap, we propose \textsc{StressEval}, a \textit{failure-driven} data synthesis framework. Rather than generating difficulty from scratch, \textsc{StressEval} uses a three-stage pipeline that turns model's failure cases into follow-up instances that remain answerable and controlable, while systematically stressing the same underlying weakness (see the right of Figure~\ref{fig:example}).
First, \textbf{Structured Error Analysis} takes instances where a model produces incorrect responses as input and employs an LLM-based analyzer to reconstruct the model’s reasoning trajectory, pinpoint the bottleneck step, and diagnose the underlying root cause. It outputs a semi-structured \emph{difficulty card} that records both the failing reasoning step and the triggering input property.
Second, guided by the difficulty card, \textbf{Dual-Perspective Instance Synthesis} leverages LLMs to generate new challenges from two complementary perspectives: for \emph{knowledge stress}, it freezes the original knowledge source, canonicalizes the missing fact as an atomic \textit{knowledge black box}, and synthesizes dependency-aware questions and answers that still hinge on that missing fact; for \emph{reasoning stress}, it synthesizes a fictitious but internally consistent virtual knowledge source with synthetic entities and then generates new question-answer pairs via a reasoning-skeleton procedure that inherits the same bottleneck and trigger while keeping the answer uniquely grounded. 
Third, a \textbf{Multi-criteria Gating} mechanism applies two LLM-based screening schemes, retaining only grounded instances that preserve the targeted difficulty card. 
We seed \textsc{StressEval} with multiple knowledge‑intensive reasoning datasets and use it to produce a focused suite of high‑difficulty evaluation items, \textsc{StressEval‑Bench}. Compared to the seed benchmarks, \textsc{StressEval‑Bench} yields substantially larger performance degradations across several state‑of‑the‑art LLMs, while preserving explicit grounding and controllable difficulty factors. 
In summary, our contributions include:
\begin{itemize}
    \item We propose \textsc{StressEval}, the first failure-driven framework for benchmarking knowledge-intensive reasoning that systematically converts observed model failures into new, difficulty-controllable test instances.
    \item We propose a complementary dual-perspective instance-synthesis method that systematically targets LLMs’ knowledge gaps and reasoning breakdowns, enabling faithful reconstruction of specific difficulty factors.
    \item Comprehensive experiments on our released dynamic \textsc{Dynamic-OneEval} demonstrate its difficulty and quality, and reveal that even state-of-the-art LLMs exhibit pronounced weaknesses in knowledge-intensive reasoning as well as fine-grained failure modes.
\end{itemize}

\section{Problem Formulation}
Assume a seed dataset $\Omega=\{(\mathcal{Q}_i,\mathcal{S}_i,\mathcal{A}_i)\}_{i=1}^{N}$, where $\mathcal{Q}_i$ is a question, $\mathcal{S}_i$ is its associated knowledge source (e.g., unstructured documents, semi-structured tables, or structured knowledge graphs), and $\mathcal{A}_i$ is the gold answer. We evaluate an LLM $\mathcal{M}$ on $\Omega$ and collect the instances on which it produces incorrect predictions. This yields a set of failure cases
$
\mathcal{E}=\{(\mathcal{Q}_i,\mathcal{S}_i,\mathcal{A}_i,\mathcal{Y}_i)\}_{i=1}^{M},
$
where $\mathcal{Y}_i$ denotes the model output and $M$ is the number of errors.
Crucially, the source context $\mathcal{S}$ may be incomplete, i.e., it may not contain all information required to answer $\mathcal{Q}$, reflecting realistic retrieval-augmented generation settings in which models must reason under partial evidence.

Our objective is to leverage $\mathcal{E}$ to automatically construct a new dataset $\Omega^* = \{(\mathcal{Q}^*_k, \mathcal{S}^*_k, \mathcal{A}^*_k)\}_{k=1}^{K}$, where each synthesized instance is derived from one failure cases and explicitly preserves the key difficulty factors that caused $\mathcal{M}$ to fail (e.g., missing or partial fact in $\mathcal{S}$, multi-hop reasoning, compositional constraints, or distractors). The resulting dataset $\Omega'$ is designed to be challenging for the model $\mathcal{M}$.

\section{\textsc{StressEval}}
\label{sec:method}
\begin{figure*}
\centering
	\includegraphics[width=\textwidth]{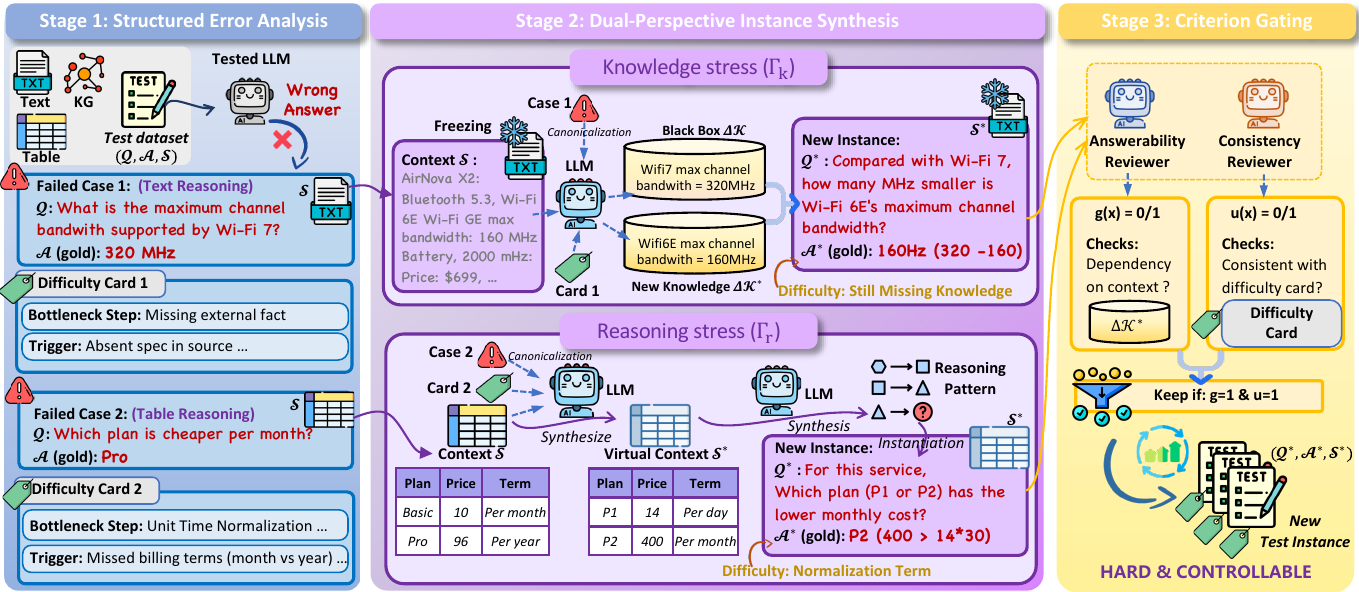}
	\caption{\textsc{StressEval} framwork. Stage 1 performs structured error analysis and produces per-case difficulty cards; Stage 2 synthesizes new instances via knowledge black-box for $\Gamma_k$ and reasoning-skeleton for $\Gamma_r$, showing original and synthesized question-answer pairs; Stage 3 applies a LLM gating mechanism to keep answerable, unambiguous, yet unsolved hard instances.} \label{fig:framework}
\end{figure*}

Figure~\ref{fig:framework} shows the overview of our proposed \textsc{StressEval} framework, which consists of: 
(i) \textbf{Structured Error Analysis} localizes the broken reasoning step and attributes the failure to a root cause; (ii) \textbf{Dual-Perspective Instance Synthesis} generates new challenging instances from the perspectives of \emph{knowledge stress} and \emph{reasoning stress}; and (iii) \textbf{Multi-criterion Gating} uses two LLM reviewers to enforce quality controls, retaining only instances that are well-grounded, and that instantiate the intended difficulty.

\subsection{Structured Error Analysis}
\label{sec:error-analysis}

For each failure case $e_i=(\mathcal{Q}_i,\mathcal{S}_i,\mathcal{A}_i,\mathcal{Y}_i)\in\mathcal{E},$ we use an LLM as an analyzer to generate a structured error report. Concretely, let $f_{\text{ana}}$ denote a reasoning LLM and  $P_{\text{ana}}$ be an analysis prompt template. We obtain
\begin{equation}
    (\gamma_i, \Psi_i)\sim p_{f_{\text{ana}}}\!\left(\cdot\ \middle|\ P_{\text{ana}}(\mathcal{Q}_i,\mathcal{S}_i,\mathcal{A}_i,\mathcal{Y}_i)\right),
\end{equation}
where $\gamma_i \in \Gamma$ is a root-cause label that characterizes the error type with a concise, precise phrase (e.g., \texttt{Entity Linking Confusion}), and $\Psi_i$ is our \emph{difficulty card} that summarizes the primary challenge underlying the failure.

\paragraph{Difficulty card.}
Our defined $\Psi_i$ aims to make each $e_i \in \mathcal{E}$ both interpretable (clarifying \emph{why} it is difficult) and controllable (specifying \emph{how} to preserve the same difficulty). It comprises the following fixed slots:
(i) \textit{Bottleneck Step:} the reasoning stage where the model fails (e.g., entity/relation recognition or comparison). For example, in the lower-left corner of Figure~\ref{fig:framework}, for the question \textit{Which plan is cheaper per month?}, the model breaks at \textit{unit-time normalization} due to misinterpreting the \textit{term} field in the table.
(ii) \textit{Trigger:} the input property, described in natural language, that triggers the failure.
In the running example, \textit{mixed billing terms} (monthly vs.\ yearly) create a unit mismatch that the model does not reconcile, causing it to compare prices without converting them to a common time basis (e.g., per month).


\subsection{Dual-Perspective Instance Synthesis}
\label{sec:synthesis}
Unlike existing methods~\cite{lin2024wildbench,zhao2025auto,wang2025rankings}, we observe that failure-driven instance synthesis is prone to \emph{root-cause entanglement}: without an explicit separation of failure sources, a synthesized instance can inadvertently mix missing-fact and reasoning breakdowns, causing the stress type to be mislabeled or the difficulty to drift (e.g., becoming trivial or unanswerable). 
This reduces controllability and weakens coverage of failure modes. 

To address this, we partition the observed failure cases $\mathcal{E}$ into two complementary perspectives according to the root-cause label $\gamma_i$ and difficulty card:
\begin{itemize}
    \item \textbf{Knowledge-stress} ($\Gamma_k$) failures: instances in which answering correctly requires information that is absent from the provided $\mathcal{S}_i$ and is also not recoverable from the model's parametric memory, resulting in errors driven by missing fact (i.e., Case 1 in Figure~\ref{fig:framework}).
    \item \textbf{Reasoning-stress} ($\Gamma_r$) failures: instances in which the provided knowledge source $\mathcal{S}_i$ contains sufficient information to answer correctly, yet the model still produces an incorrect response due to misreading the evidence or flawed reasoning (i.e., Case 2 in Figure~\ref{fig:framework}).

\end{itemize}
Below, we will introduce how to synthesize new, high-quality, and challenging test instances from each perspective.

\subsubsection{Knowledge-stress Instance Synthesis}
\label{subsec:typek}

This branch aims to preserve the knowledge stress by \emph{freezing the original knowledge context} $\mathcal{S}_i$. Intuitively, since the model answered question $\mathcal{Q}_i$ incorrectly due to missing fact, it is reasonable to expect that it will also answer incorrectly any questions derived from $\mathcal{Q}_i$.
Specifically, for any failure case $e_i$ with $\gamma_i\in\Gamma_k$, we keep:
$
\mathcal{S}^*_i \coloneqq \mathcal{S}_i.
$

\paragraph{Knowledge black box.}
For knowledge-stress failures, the missing fact is external to the provided source $\mathcal{S}_i$, so a faithful step-by-step decomposition is often underspecified: the model fails not because of an explicit reasoning error over $\mathcal{S}_i$, but because a crucial fact is unavailable. To preserve this \textit{knowledge gap} while enabling controlled synthesis, we collapse the original question–answer pair into a self-contained atomic knowledge statement (like \textit{black box}), denoted as $K_0$, and treat it as an indivisible unit that can be re-queried or rephrased without altering the underlying missing fact:
\begin{equation}
K_0 \sim p_{\mathcal{L}_{\text{can}}}\!\left(\cdot\ \middle|\ P_{\text{can}}(\mathcal{Q}_i,\mathcal{A}_i)\right),
\end{equation}
where $\mathcal{L}_{\text{can}}$ is an reasoning LLM used for canonicalization (with prompt template $P_{\text{can}}$), and $P_{\text{can}}$ is a prompt template that converts the pair $\left(\mathcal{Q}_i,\mathcal{A}_i\right)$ into a declarative, self-contained knowledge statement.
For instance, from $\mathcal{Q}_i$: \textit{What is the maximum channel bandwidth supported by Wi-Fi 7?} and $\mathcal{A}_i$: \textit{320\,MHz}, we obtain canonical $K_i$: \textit{The maximum channel bandwidth supported by Wi-Fi 7 is 320\,MHz.}

\paragraph{Dependency-aware QA synthesis.}
Then, we extract a new, definitive piece of knowledge $\Delta K$ from $\mathcal{S}_i$ (i.e., directly supported by $\mathcal{S}_i$ and thus reliable), and combine it with the knowledge black box $K_0$ to synthesize a new question–answer pair $\left(\mathcal{Q}^*_i,\mathcal{A}^*_i\right)$. The resulting instance remains faithful to the fixed source $\mathcal{S}^*_i$, while ensuring that correctly solving it still hinges on first resolving the original $\mathcal{Q}_i$.
\begin{equation}
\Delta K \sim p_{f_{\text{ext}}}\!\left(
    \cdot \ \middle|\ P_{\text{ext}}(\mathcal{S}^*_i,\Psi_i)
\right),
\end{equation}
\begin{equation}
\left(\mathcal{Q}^*_i,\mathcal{A}^*_i\right) 
\sim p_{f_{\text{know}}}\!\left(
    \cdot \ \middle|\ 
    P_{\text{know}}\!\left(\mathcal{S}^*_i,\Psi_i, K_0,\Delta K\right)
\right),
\end{equation}
where $f_{\text{ext}}$ is an LLM-based knowledge extractor (with prompt template $P_{\text{ext}}$) that identifies a context-grounded fact $\Delta K$ from $\mathcal{S}_i^*$, and $f_{\text{know}}$ (with template $P_{\text{know}}$) generates a new pair $\left(\mathcal{Q}^*_i,\mathcal{A}^*_i\right)$ conditioned on $\mathcal{S}_i$, the difficulty card $\Psi_i$, the missing-fact $K_0$, and the extracted fact $\Delta K$.

\subsubsection{Reasoning-stress Instance Synthesis}
\label{subsec:typer}

This branch aims to reproduce the same \emph{reasoning trap} in new instances that are fully answerable from the provided context. For any failure case $e_i$ with $\gamma_i\in\Gamma_r$, we create a \emph{virtual} knowledge source with fictitious entities to prevent the model from exploiting parametric memory.

\paragraph{Virtual knowledge source synthesis.}
We first employ an LLM to sample a synthetic universe
\begin{equation}
    \mathcal{U}_i=(\mathbf{E},\mathbf{R},\mathbf{V})\sim
p_{f_{U}}\!\left(\cdot\ \middle|\ P_U(\mathcal{Q}_i, \mathcal{S}_i, \mathcal{A}_i, \Psi_i)\right),
\end{equation}
where $\mathbf{E}$ is a set of fictitious entity identifiers, $\mathbf{R}$ defines a relation/field schema, and $\mathbf{V}$ assigns attribute values (e.g., numbers, timestamps, categories) along with unit and formatting conventions. We then render $\mathcal{U}$ into a concrete but virtual knowledge source $\mathcal{S}^*_i$ based on the original $\mathcal{S}_i$:
\begin{equation}
    \mathcal{S}^*_i \sim p_{f_{S}}\!\left(\cdot\ \middle|\ P_S(\mathcal{U}_i, \mathcal{Q}_i, \mathcal{S}_i, \mathcal{A}_i, \Psi_i)\right).
\end{equation}
We enforce a \emph{synthetic-entity policy}: all named entities are sampled from a reserved namespace (e.g., \texttt{P1} and \texttt{P2} in Figure~\ref{fig:framework}) to minimize overlap with real-world facts. 

\paragraph{QA pair synthesis with pattern inheritance.}
Given the virtual $\mathcal{S}^r_i$ and the difficulty card $\Psi_i$, we synthesize a new question-answer pair while explicitly \emph{inheriting} the original reasoning bottleneck and trigger.
Here, we use a two-step LLM procedure: (i) generate a \emph{reasoning skeleton} that operationalizes $\Psi_i$ into a minimal sequence of required reasoning operations and a concrete \emph{trap injection} (e.g., an ambiguity/distractor that targets the bottleneck step), and (ii) render the skeleton into a natural-language question whose gold answer is uniquely derivable from $\mathcal{S}^*_i$.

Rather than asking an LLM to directly generate a new question, we first generate an explicit \emph{reasoning skeleton} $\mathcal{G}_i$ that operationalizes $\Psi_i$ into minimal, checkable components:
required evidence fields/records in $\mathcal{S}^*_i$,
the bottleneck operation to be executed (as stated in $\Psi_i$),
a \emph{trap injection} consistent with the trigger (a distractor that yields a plausible wrong answer if the bottleneck is skipped),
and a short gold reasoning trace that points to concrete evidence.
Formally,
\begin{equation}
    \mathcal{G}_i \sim p_{f_{\text{skel}}}\!\left(\cdot\ \middle|\ P_{\text{skel}}(\mathcal{S}^*_i,\Psi_i)\right),
\end{equation}
\begin{equation}
    (\mathcal{Q}^*_i,\mathcal{A}^*_i)\sim
p_{f_{\text{reason}}}\!\left(\cdot\ \middle|\ P_{\text{reason}}(\mathcal{S}^*_i,\Psi_i,\mathcal{G}_i)\right).
\end{equation}
We constrain the LLM to (a) mention only entities/fields appearing in $\mathcal{S}^*_i$, (b) include at least one distractor consistent with the trigger in $\Psi_i$ such that skipping the bottleneck step leads to a plausible but incorrect answer, and (c) ensure $\mathcal{A}^*_i$ is supported by an explicit reasoning trace grounded in $\mathcal{S}^*_i$.

\subsection{Multi-criterion Gating}
\label{sec:filter}
In this stage, we perform \emph{multi-criterion gating} to control the quality of synthesized instances. For each candidate instance $x_i=(\mathcal{S}_i,\mathcal{Q}_i,\mathcal{A}_i,\Psi_i),$ we query two LLM-based reviewers, each prompted to return a structured verdict together with a calibrated confidence score. Specifically, we use: (1) an \emph{Answerability} reviewer, which verifies that the question is answerable under the intended stress type and that the gold answer is properly grounded---i.e., supported by explicit spans/records in $\mathcal{S}_i$ for reasoning-stress, or demonstrably requiring a missing external fact beyond $\mathcal{S}_i$ for knowledge-stress; and (2) a \emph{Consistency} reviewer, which attempts to solve the instance, checks internal consistency among $\mathcal{S}_i,\mathcal{Q}_i,\mathcal{A}_i,$ and verifies that the difficulty card $\Psi_i$ is actually instantiated rather than merely stated.

Each reviewer outputs a binary decision with confidence. We then aggregate their outputs into two binary gating signals: $g_i\in\{0,1\}$ indicating answerability, and $u_i\in\{0,1\}$ indicating consistency, $\text{keep}(x_i)=\mathbb{I}\big[g_i=1\big]\cdot\mathbb{I}\big[u_i=1\big]$,
Notably, for knowledge-stress instances, grounding instead verifies the \emph{dependency} on the missing-fact black box (i.e., solving $\mathcal{Q}_i$ necessarily requires resolving $\Delta K$ in addition to any context-grounded fact), and this dependency check is incorporated into $g_i$.


\section{\textsc{Dynamic-OneEval}}
\subsection{Seed Data \& Implementation Details}
To initialize \textsc{Dynamic-OneEval}, we draw seed instances from widely used benchmarks.
\textbf{a) Text Reasoning:}
\textit{HotpotQA}~\cite{yang2018hotpotqa} is a Wikipedia-based multi-hop QA benchmark that requires combining evidence across multiple documents to answer. 
\textbf{b) KG Reasoning:}
\textit{ComplexWebQ}~\cite{talmor2018web} is a compositional KBQA benchmark designed to answer complex questions by executing multi-relation queries over a knowledge graph. 
\textbf{c) Table Reasoning:}
\textit{WikiTableQuestion} (WTQ)~\cite{pasupat2015compositional}
\textit{TabFact} is a table-based fact verification dataset that asks whether a natural-language statement is entailed or contradicted by a given Wikipedia table. 

We generated seed failure cases $\mathcal{E}$ by prompting GPT-5.2~\cite{openai_api_2025}, a strong LLM , to produce \textit{chain-of-thought}~\cite{wei2022chain} answers and then verifying its outputs to collect instances where the model erred. For end-to-end quality control, the strong GPT-5.2 model was used throughout the StressEval pipeline for $f_\text{ana}$, $f_\text{can}$, $f_\text{ext}$, $f_\text{know}$, $f_\text{U}$, $f_\text{U}$. All API calls used deterministic decoding with temperature was set to $0.0$, top-p was set to $0.95$, and a maximum token limit was set to $2048$. For each bad case $e \in \mathcal{E}$, StressEval will synthesize 5 $\sim$ 10 instances. 
In addition, we do not synthesize knowledge-stress instances for table reasoning, since the case of the WTQ dataset already contain all evidence required to answer the questions. Detailed prompts are listed in Appendix~\ref{sec:prompt}.

\subsection{Benchmark Statistic}
Table~\ref{tab:overall_stats} reports statistics of the constructed dataset. For each split, we provide the number of synthesized instances $|\Omega^*|$, the size of the seed dataset $|\Omega|$, the size of the failure-case set $|\mathcal{E}|$, the average context and question lengths, the mean token number of $\mathcal{S}^*$ and $\mathcal{Q}^*$ (reported as $\mathrm{avg}\,|\mathcal{S}^*|$ and $\mathrm{avg}\,|\mathcal{Q}^*|$), the number of root causes $|\Gamma|$, and the token-level similarity between the synthesized and original questions, $\mathrm{Sim}(\mathcal{Q}, \mathcal{Q}^*)$.
Overall, \textsc{StressEval} generates a large number of challenging instances $|\Omega|$ from a relatively small collection of failure cases $|\mathcal{E}|$, demonstrating strong leverage of limited supervision. Moreover, in principle, with a continual stream of new failure cases, the construction process can keep producing arbitrarily many difficult instances, i.e., the dataset size is unbounded as the failure-case pool grows.
More instances of \textsc{Dynamic-OneEval} are shown in Appendix~\ref{sec:instance}.

\begin{table*}[t]
\caption{Overall statistics of \textsc{Dynamic-OneEval} and the results of human evaluation of different splits.}
\centering
\small
\setlength{\tabcolsep}{6pt}
\begin{tabular}{llcccccccccc}
\toprule
\textbf{Type} &\textbf{Split} & $|\Omega^*|$ & $|\Omega|$ & $|\mathcal{E}|$ & \textbf{Avg} $|\mathcal{S}^*|$ & \textbf{Avg} $|\mathcal{Q}^*|$ & $|\Gamma|$ &Sim($\mathcal{Q}$, $\mathcal{Q}^*$) & \textbf{A(\%)} & \textbf{U(\%)} & \textbf{F(\%)} \\
\cmidrule(lr){1-1} \cmidrule(lr){2-2} \cmidrule(lr){3-5} \cmidrule(lr){6-7} \cmidrule(lr){8-9} \cmidrule(lr){10-12} 
\multirow{2}[1]{*}{\textbf{Text}} &K-stress $\Gamma_k$  & 468 & 1000 & 78 & 1047.6 & 30.4 & 2 & 52.6 &100.0 &100.0 &96.0   \\
&R-stress $\Gamma_r$ & 388 & 500 & 146 & 407.3 & 30.1 & 10 & 53.5 &98.0 &100.0 &90.0  \\
\cmidrule(lr){1-1} \cmidrule(lr){2-2} \cmidrule(lr){3-5} \cmidrule(lr){6-7} \cmidrule(lr){8-9} \cmidrule(lr){10-12} 
\multirow{2}[1]{*}{\textbf{KG}} &K-stress $\Gamma_k$  & 432 &400 & 114 & 919.8 & 23.1 &3 &50.3 & 100.0 &100.0 &98.0  \\
&R-stress $\Gamma_r$ & 347 & 1200 & 64 & 708.5 & 13.8 &7 &48.5 & 94.0 &100.0  &88.0\\
\cmidrule(lr){1-1} \cmidrule(lr){2-2} \cmidrule(lr){3-5} \cmidrule(lr){6-7} \cmidrule(lr){8-9} \cmidrule(lr){10-12} 
\multirow{1}[1]{*}{\textbf{Table}}
&R-stress $\Gamma_r$ & 294 & 1000 & 47 & 236.6 & 20.9 & 8 & 51.0 &100.0 &98.0 &96.0   \\
\cmidrule(lr){1-2} \cmidrule(lr){3-5} \cmidrule(lr){6-7} \cmidrule(lr){8-9} \cmidrule(lr){10-12} 
\multicolumn{2}{c}{\textbf{\textsc{Dynamic-OneEval}}} &1929 &4100  &449  &705.6  &24.3 &30 &51.3 &98.5 &99.7 &93.8   \\
\bottomrule
\end{tabular}
\label{tab:overall_stats}
\end{table*}

\subsection{Diversity of Root Cases}
\begin{figure}
\centering
	\includegraphics[width=0.48\textwidth]{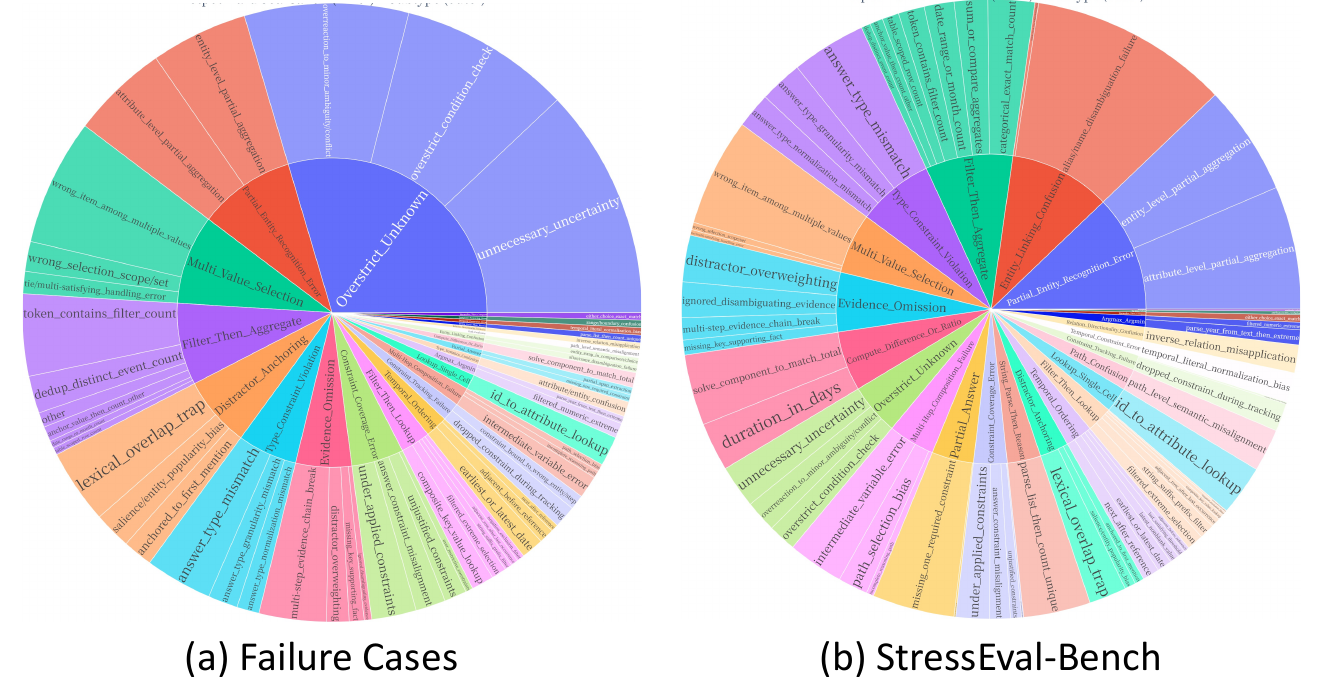}
	\caption{Left: root-cause distribution among the seed failure cases $\mathcal{E}$. Right: root-cause distribution of our \textsc{Dynamic-OneEval}, which is comparatively more balanced.} \label{fig:pie}
\end{figure}

Figure~\ref{fig:pie} shows the distribution of root-cause labels $\gamma \in \Gamma$ for both the original failure set $\mathcal{E}$ and the synthesized \textsc{Dynamic-OneEval}. The inner ring indicates coarse-grained root-cause groups, while the outer ring breaks them down into the corresponding fine-grained categories.
Overall, the synthesized benchmark largely preserves the diversity of error types present in the collected failures, and it also introduces modest expansion into additional categories. Notably, the label distribution in \textsc{Dynamic-OneEval} is more balanced than in the seed failure set, reducing skew toward a few dominant root causes and yielding broader coverage of fine-grained model weaknesses. Due to space constraints, we provide a clearer version of the figures in Appendix~\ref{sec:distribution}.

\subsection{Human Evaluation}
\label{sec:human-eval}

We conducted a lightweight human evaluation to assess the quality of \textsc{Dynamic-OneEval}. We randomly sampled 50 instances spanning both knowledge-stress $\Gamma_k$ and reasoning-stress $\Gamma_r$, covering diverse root-cause labels $\gamma \in \Gamma$, and asked three annotators to provide judgments along three dimensions. First, \emph{answerability} (\textbf{A}) evaluates whether the instance can be resolved from the provided context: for $\Gamma_r$, whether the gold answer $\mathcal{A}$ is supported by explicit context in $\mathcal{S}$; for $\Gamma_k$, whether solving the question $\mathcal{Q}$ necessarily requires an external fact missing from $\mathcal{S}$. Second, \emph{unambiguity} (\textbf{U}) assesses whether a unique gold answer exists. Third, \emph{faithfulness to difficulty cards} (\textbf{F}) measures whether the difficulty card $\Psi$ is correctly instantiated, meaning that the specified bottleneck step and trigger are present in the instance. 

The results are reported in the right panel of Table~\ref{tab:overall_stats}. Overall, human judgments show consistently high \textbf{A} and \textbf{U} across splits, indicating that instances are generally solvable from the provided context and admit a unique gold answer. In contrast, \textbf{F} exhibits larger variance, with reasoning-stress splits showing notably lower faithfulness (e.g., $88.0$\% $\sim$ $96.0$\%) than knowledge-stress splits (up to $98.0$\%), suggesting that mismatches more often arise from imperfect instantiation of the intended bottleneck and trigger rather than from missing fact or ambiguity. Aggregated over the benchmark, the strong overall scores support the reliability of our framework.

\section{Experiments}

\subsection{Experimental Setup}

We evaluate a set of representative open-source and proprietary LLMs on \textsc{Dynamic-OneEval}, covering the Llama3.1 series~\cite{grattafiori2024llama} (Llama3.1-8B and Llama3.1-70B), the Qwen family~\cite{yang2024qwen2,yang2025qwen3} (Qwen2.5-72B and Qwen3-235B) and QWQ-32B~\cite{qwq32b}, as well as proprietary models including GPT-5.2~\cite{openai_api_2025}, Gemini3-pro~\cite{google_gemini_3_pro_models_doc}, DeepSeek-V3.2~\cite{liu2025deepseek}, Claude-Sonnet-4.5~\cite{anthropic_claude_sonnet_4_5_news}, and Doubao-Seed-1.6~\cite{bytedance_seed1_6_page}.   
To ensure a fair comparison, we apply a unified prompting template across all models.  
We use exact-match accuracy as the primary metric, reporting results by knowledge type (Text, KG, and Table) and stress type (K-Stress/R-Stress), together with per-type averages and an overall aggregated score.

\subsection{Main Results}

Table~\ref{tab:main_results} reports the performance of different LLMs on  \textsc{Dynamic-OneEval}. The results show that \textsc{Dynamic-OneEval} is genuinely hard and far from saturated: even the strongest proprietary model (Gemini3-pro) reaches only $48.2\%$ overall, and the best open-source model (QWQ-32B) achieves $39.8\%$, with substantial error persisting across all knowledge types. This indicates that our failure-driven generation goes beyond superficial adversarial trickery, producing instances that are both challenging and unambiguous, and that reliably surface brittleness even in frontier LLMs, particularly on knowledge-intensive failure modes.

A consistent diagnostic pattern is that K-Stress is the dominant bottleneck, most severely in text reasoning: open-source models remain near-floor on Text K-Stress despite noticeably higher Text R-Stress, and although proprietary models improve (Gemini3-pro achieves $29.8$\%), most still hover around $10$ on text K-Stress. In KG reasoning, R-Stress is comparatively high and clustered, while K-Stress drops sharply, indicating that graphs can skeleton compositional reasoning but cannot compensate when crucial knowledge is stressed or withheld. Finally, table reasoning yields high R-Stress for several models (e.g., Claude-Sonnet-4.5 achieves $80.9$\%, Gemini3-pro achieves $75.5$\%), yet overall scores remain limited because knowledge-stress failures persist, reinforcing \textsc{Dynamic-OneEval} as a targeted, controlable stress test rather than a knowledge type-specific formatting challenge.

\begin{table*}
\centering
\caption{Performance (\%) of open-source and proprietary LLMs on \textsc{Dynamic-OneEval}. All test instances were generated using GPT-5.2.}
\label{tab:main_results}
\scalebox{0.88}{
\begin{tabular}{c l c c c c c c c c c c}
\toprule
\multirow{2}[1]{*}{\textbf{\makecell{Type}}} & \multirow{2}[1]{*}{\textbf{Model}} & \multicolumn{3}{c}{\textbf{Text}} & \multicolumn{3}{c}{\textbf{KG}} & \multicolumn{1}{c}{\textbf{Table}} & \multirow{2}[1]{*}{\textbf{Overall}} \\
\cmidrule(lr){3-5} \cmidrule(lr){6-8} \cmidrule(lr){9-9} 
& & K-Stress & R-Stress & Avg. & K-Stress & R-Stress & Avg. & R-Stress \\
\cmidrule(lr){1-1} \cmidrule(lr){2-2} \cmidrule(lr){3-5} \cmidrule(lr){6-8} \cmidrule(lr){9-9} \cmidrule(lr){10-10}
\multirow{5}[1]{*}{\textit{\makecell{Open-sourced\\LLM}}} &Llama3.1-8B~\cite{grattafiori2024llama} & 15.6 & 20.1 & 17.6 & 11.2 & 44.1 & 25.9 & 42.2 & 24.7 \\
&Llama3.1-70B~\cite{grattafiori2024llama} & \textbf{21.4} & 21.1 & 21.3 & 27.8 & 51.3 & 38.3 & 44.9 & 31.7 \\
&Qwen2.5-72B~\cite{yang2024qwen2} & 14.7 & 21.6 & 17.8 & \textbf{33.9} & \textbf{53.0} & \textbf{42.4} & 58.5 & 33.9  \\
&QWQ-32B~\cite{qwq32b} & 17.9 & \textbf{23.7} & \textbf{20.5} & 22.4 & 50.7 & 35.0 & \textbf{74.8} & \textbf{34.6}  \\
&Qwen3-235B~\cite{yang2025qwen3} & 16.7 & 21.6 & 18.9 & 13.0 & 50.7 & 29.8 & 54.1 & 28.7  \\
\cmidrule(lr){1-1} \cmidrule(lr){2-2} \cmidrule(lr){3-5} \cmidrule(lr){6-8} \cmidrule(lr){9-9}  \cmidrule(lr){10-10}
\multirow{5}[1]{*}{\textit{\makecell{Proprietary\\LLM}}} &GPT-5.2~\cite{openai_api_2025} & 26.9 & 26.8 & 26.9 & 10.0 & 49.6 & 27.6 & 73.8 & 34.3  \\
&Gemini3-pro~\cite{google_gemini_3_pro_models_doc} & \textbf{54.3} & \textbf{27.6} & \textbf{42.2} & \textbf{28.6} & 54.2 & \textbf{40.0} & 75.5 & \textbf{46.4}  \\
&DeepSeek-V3.2~\cite{liu2025deepseek} & 15.6 & 13.9 & 14.8 & 15.7 & \textbf{55.0} & 33.2 & 50.3 & 27.7  \\
&Claude-Sonnet-4.5~\cite{anthropic_claude_sonnet_4_5_news} & 31.2 & 23.7 & 27.8 & 13.9 & 53.4 & 31.5 & \textbf{80.9} & 37.4  \\
&Doubao-Seed-1.6~\cite{bytedance_seed1_6_page} & 32.9 & 15.7 & 25.1 & 1.9 & 51.8 & 24.1 & 47.6 & 28.1  \\
\bottomrule
\end{tabular}
}
\end{table*}

\subsection{Ablation Studies}
\label{sec:ablation}
We fix GPT-5.2 as the synthesis backbone, generate instances under each ablated setting, and evaluate the resulting datasets using DeepSeek V3.2, together with human judgments on the corresponding instances.
Here, we consider three ablation groups. \textbf{Pipeline-level} ablations remove either (a) \emph{error analysis}, to test whether bottleneck identification and trigger localization are necessary for controllable synthesis, or (b) \emph{criterion gating}, to quantify its role in enforcing grounding, answerability, unambiguity, and difficulty preservation. \textbf{Knowledge-stress} ablations remove either (c) the \emph{missing-fact black box} $\Delta K$, to test whether explicitly modeling atomic missing facts stabilizes reliance on external knowledge, or (d) \emph{context freezing}, allowing $\mathcal{S}$ to be rewritten and potentially filling the intended knowledge gap. \textbf{Reasoning-stress} ablations remove either (e) \emph{virtual context} $\mathcal{S}^*$ construction, to assess whether entity virtualization prevents parametric-memory shortcuts, or (f) the \emph{reasoning skeleton}, to test whether explicitly instantiating the bottleneck and injecting traps improves faithfulness and grounding.

Table~\ref{tab:ablation} reports the ablation results for \textsc{StressEval}, showing that its components must operate in concert to preserve both difficulty and validity. At the pipeline level, removing error analysis substantially degrades performance (e.g., DeepSeek V3.2 drops from $27.7\%$ overall in Table~\ref{tab:main_results}) and sharply reduces human-rated faithfulness, indicating that bottleneck and trigger are crucial for controllable synthesis. Removing criterion gating also consistently lowers human faithfulness, confirming its role in enforcing grounding, unambiguity, and difficulty preservation.
For K-stress, disabling the missing-fact black box $K$ produces the most severe failures, while unfreezing $\mathcal{S}$ weakens the intended knowledge gap, consistent with context leakage. For R-stress, removing the virtual context $\mathcal{S}^*$ or the two-step reasoning-skeleton procedure yields instances that are noticeably easier and less faithful, supporting the claim that virtualization mitigates parametric-memory shortcuts and that an explicit skeleton enforces context-grounded reasoning.

\begin{table}[t]
\centering
\caption{Ablation study on \textsc{Dynamic-OneEval} (using GPT-5.2 for data synthesis). We report performance (\%) of DeepSeek V3.2.}
\label{tab:ablation}
\scalebox{0.86}{
\begin{tabular}{l c c c c c}
\toprule
\textbf{Variant} & \textbf{Text}$\downarrow$ & \textbf{KG}$\downarrow$ & \textbf{Table}$\downarrow$ & \textbf{A} & \textbf{F} \\
\midrule
\textsc{Dynamic-OneEval} & 14.8 & \textbf{33.2}  & 50.3
&\textbf{98.4} &\textbf{93.6}  \\
\midrule
w/o error analysis & 68.9 & 40.3  & 61.7 & 90.0 & 48.0  \\
w/o criterion gating & \textbf{11.7} & 43.1 & \textbf{39.2} & 92.0 & 84.0  \\
\midrule
Knowledge-stress & \textbf{15.6} & \textbf{15.7} & -- & \textbf{100.0}  & \textbf{97.0 } \\
\midrule
w/o black box  & 66.7 & 83.6  & -- & 98.0 & 96.0  \\
w/o freezing $\mathcal{S}$  & 32.1 & 23.8  & -- & 84.0 & 92.0  \\
\midrule
Reasoning-stress & \textbf{13.9} & \textbf{55.0}  & 50.3 & 97.3 &\textbf{91.3}  \\
\midrule
w/o virtual $\mathcal{S}^*$  & 59.7 & 61.8  & 54.5 & 96.0 & 88.0  \\
w/o reasoning skeleton  & 56.8 & 66.5  & \textbf{34.1} & \textbf{98.0}  &80.0  \\
\bottomrule
\end{tabular}
}
\end{table}

\subsection{Performance on Different Root Causes}
\begin{figure*}
\centering
	\includegraphics[width=\textwidth]{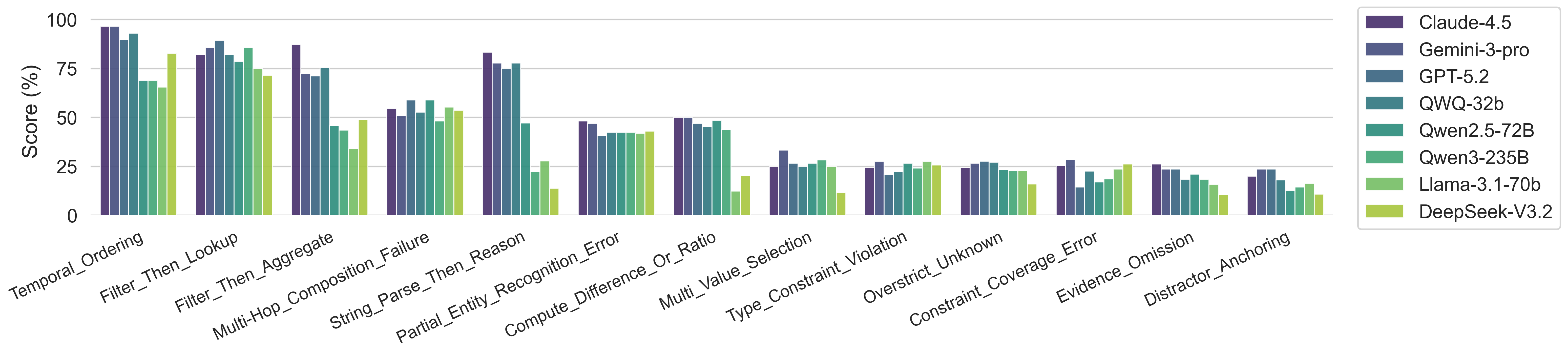}
	\caption{Performance of different LLMs on each root cause. Due to space limitations, only representative root causes are included here.} \label{fig:root_cause_score}
\end{figure*}

Figure~\ref{fig:root_cause_score} compares models across fine-grained root error causes; due to space constraints, we report representative categories. Performance is highly error-dependent: frontier proprietary models excel on structured temporal/compositional skills. For instance, Claude-Sonnet-4.5 achieves $95\%$ on \texttt{Temporal\_Ordering}, whereas gaps widen for failures requiring robust entity grounding and multi-step control. For instance, \texttt{Partial\_Entity\_Recognition\_Then\_Reason} falls from $85\%$ (Claude-4.5) to $15\%$ (DeepSeek-V3.2). Across models, \texttt{Evidence\_Omission} and \texttt{Distractor\_Anchoring} remain difficult, suggesting that brittleness stems less from surface retrieval than from evidence use under adversarial distractions.

\subsection{Impact of Different Backbones.}
\begin{figure*}
\centering
	\includegraphics[width=0.9\textwidth]{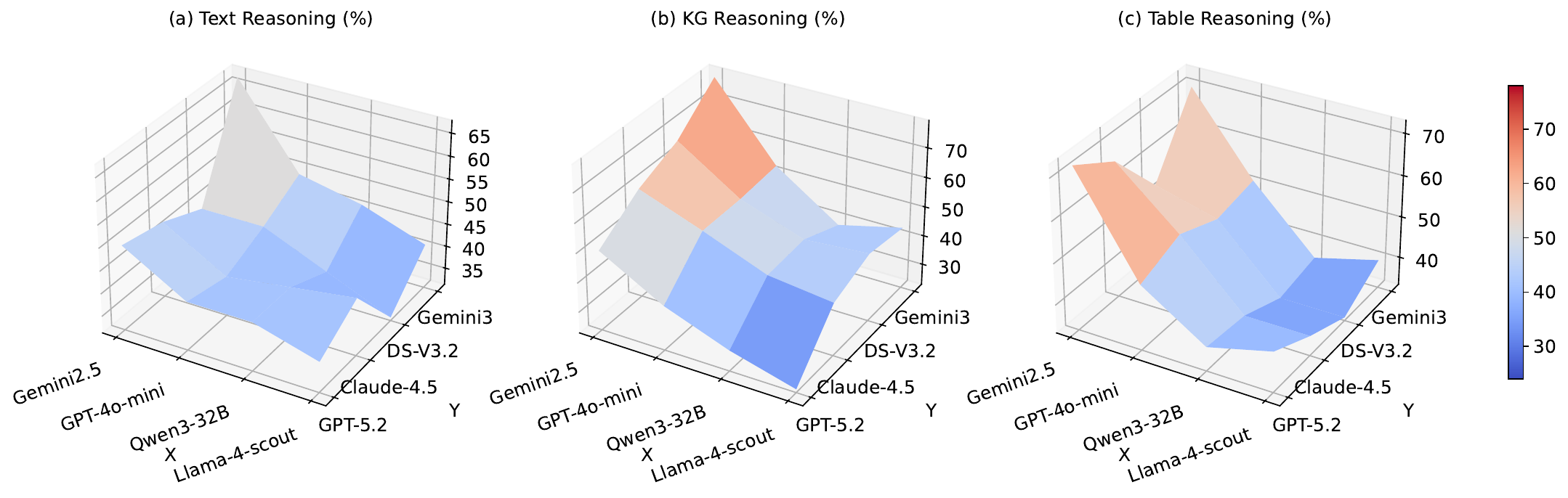}
	\caption{Performance of \textsc{StressEval} using different backbones. We use different backbone LLMs (x-axis) to run \textsc{StressEval} and generate evaluation instances, and then benchmark a set of target LLMs (y-axis) on the resulting datasets.} \label{fig:xyz}
\end{figure*}

To assess \textsc{StressEval}'s robustness to the backbone, we re-ran the pipeline with several lightweight/low-cost LLMs as generators and evaluated the resulting instances using representative LLMs. Figure~\ref{fig:xyz} summarizes the results. Model rankings are largely stable across backbones, suggesting limited generator-induced bias, while the backbone mainly affects absolute difficulty: Llama-4-scout generally yields harder cases (lower scores) and Gemini2.5 tends to yield easier ones (higher scores) across all three settings. Notably, Gemini3 performs particularly well on Gemini2.5-generated cases, consistent with potential within-family transfer effects.

\section{Related Work}

\paragraph{Knowledge-Intensive Reasoning Benchmarks}
Early success of LLMs, exemplified by models like GPT-4 \cite{openai2023gpt4} and DeepSeek~\cite{liu2024deepseek}, demonstrated impressive capabilities across a spectrum of language understanding and reasoning tasks. However, conventional static evaluation practices have struggled to keep pace with the rapid evolution of these models. Issues such as benchmark contamination~\cite{deng2024investigating,donggeneralization}, where models may memorize test data present in their vast training corpora, lead to an inflated sense of performance and obscure genuine generalization abilities \cite{lin2024rethinking}. This problem is particularly acute in knowledge-intensive settings~\cite{chen2025k,chen2022outlining}, where correctness relies on up-to-date facts, strict grounding, and robust multi-step inference, rather than superficial pattern matching \cite{jiang2024evaluating}. The need for benchmarks that can genuinely assess an LLM's ability to reason with and apply external knowledge, rather than merely recall memorized patterns, remains a significant challenge. Our work directly addresses this by creating instances that expose specific knowledge gaps and reasoning failures that are otherwise masked by existing benchmarks.

\paragraph{Dynamic LLM Benchmarks}

Prior efforts in automatic benchmarking~\cite{wang2025rankings} broadly fall into four categories: (i) static, human-curated benchmark suites; (ii) LLM-as-judge evaluation frameworks such as MT-Bench~\cite{zheng2023judging}; (iii) agent-driven, end-to-end automated arenas (e.g., Auto-Arena~\cite{zhao2025auto}); and (iv) \textit{in-the-wild} robustness benchmarks mined from real user interactions, such as WildBench~\cite{lin2024wildbench}. These approaches improve scalability and realism but have complementary drawbacks: static benchmarks saturate, risk data contamination, and offer limited diagnostics; mined \textit{in-the-wild} sets reveal real failures but lack controllable latent factors for causal stress tests~\cite{valiulla2025benchmarks}; and fully automated agent pipelines boost throughput and ranking but may generate cases that are not reliably answerable, controlable, or causally faithful—limiting debugging and improvement~\cite{shukla2025large}.
Model-in-the-loop, continual, and adversarial test-generation frameworks aim to keep evaluations up to date and to uncover brittle heuristics that static benchmarks can miss~\cite{white2024livebench,li2024forewarned}. However, without strict constraints on answerability and provenance, such dynamic generation can drift toward unanswerable, underspecified, or spurious examples, weakening both the validity and the actionability of the resulting evaluations~\cite{li2024forewarned,white2024livebench}.
Compared with these benchmarks, our \textsc{StressEval} prioritizes causal attribution, difficulty preservation, contamination resistance.

\section{Conclusion}

We present \textsc{StressEval}, a failure-driven pipeline that converts observed LLM failures into dynamic, answerable, and high-difficulty evaluation instances through structured error analysis, dual-perspective instance synthesis, and multi-criterion gating. Instantiated from knowledge-intensive seed datasets, \textsc{StressEval} produces grounded and controlable tests that induce substantially larger, and more diagnostic, performance drops across frontier models than standard static benchmarks, while preserving explicit difficulty factors to support actionable iteration.  
Future work will investigate how different combinations of LLMs within the same \textsc{StressEval} framework affect data generation, e.g., using heterogeneous generator–critic–verifier ensembles, to improve diversity, robustness, and controllability.



\bibliographystyle{named}
\bibliography{ijcai26}

\appendix
\onecolumn
\newpage

\section{Prompt used in \textsc{StressEval}}
\label{sec:prompt}

This section provides the main prompt for building \textsc{StressEval}.

\subsection{Error Analysis}

\begin{tcolorbox}[
    colback=gray!10,
    colframe=darkgray,
    title=Prompt for Text R\_stress (for tracing from badcases),
    coltitle=white,
    fonttitle=\bfseries,
    boxrule=1.5pt,
    width=\linewidth,
    before skip=1em,
    after skip=1em,
    breakable,
    left=1mm, right=1mm, boxsep=1mm
]
\begingroup
\footnotesize\ttfamily
\begin{Verbatim}[
  breaklines=true,
  breakanywhere=true,
  breaksymbolleft={},
  breaksymbolright={}
]
You are a tracing assistant for Hotpot-style multi-hop QA.

You are given:
- A question
- Context passages with titles and numbered sentences
- A model's predicted answer
- The model's short explanation and raw output (may be incomplete)

Your task:
Reconstruct a plausible STRUCTURED reasoning trace that matches the model's predicted answer,
using ONLY the provided contexts as evidence.

Output MUST be exactly ONE valid JSON object with this schema:
{
  "trace": [
    {
      "step": 1,
      "op": "<one of: bridge_entity, coreference_resolution, constraint_tracking, comparison, attribute_lookup, set_aggregation, distractor_filtering>",
      "evidence": ["<title>", <sent_id>],
      "result": "<short intermediate string>"
    },
    ...
  ]
}

Rules (CRITICAL):
1) Use ONLY the provided contexts. Do NOT use external knowledge.
2) The trace must have 2 to 4 steps.
3) Each step MUST cite exactly ONE evidence sentence using ["title", sent_id], where sent_id is 0-based.
4) Evidence must be VALID: title must exactly match a provided title; sent_id must be in range for that title.
5) The trace MUST use evidence from at least TWO DIFFERENT titles.
6) The final step's "result" MUST exactly equal the given model predicted answer.
7) Each step's "result" must be SHORT (a name/phrase/number/year/place), not a paragraph.

EVIDENCE-RESULT CONSISTENCY (MOST IMPORTANT):
8) Each step's result MUST be directly supported by its evidence sentence:
   - Prefer copying a short span that appears verbatim in the evidence sentence.
   - If minor normalization is needed (e.g., removing parentheses, quotes, or trailing punctuation), keep the surface form essentially identical.
   - Do NOT invent new entities or facts that do not appear in the cited sentence.
9) Use the op field honestly:
   - attribute_lookup: evidence states an attribute/value for an entity.
   - bridge_entity: evidence provides the link entity needed to jump to another title.
   - coreference_resolution: evidence clarifies what a pronoun/alias refers to.
   - constraint_tracking: evidence introduces or applies a constraint (year, role, membership, “only”, “first”, etc.).
   - comparison: evidence provides a comparable attribute (year/size/rank) used to compare.
   - set_aggregation: evidence provides membership of a set (list, group, categories).
   - distractor_filtering: evidence is tempting but not the true support; use it to reflect the model’s likely anchoring.

10) If the predicted answer cannot be supported by any evidence sentence without inventing facts:
    - Still output a trace that matches the predicted answer, but make it reflect a likely failure:
      use distractor_filtering for at least one step where the evidence is related but insufficient,
      and keep results as short spans present in the evidence when possible.

Output constraints (STRICT):
- Output ONLY the JSON object, no markdown, no extra text.
\end{Verbatim}
\endgroup
\end{tcolorbox}
\begin{tcolorbox}[
    colback=gray!10,
    colframe=darkgray,
    title=Prompt for Text R\_stress (for generating error reports),
    coltitle=white,
    fonttitle=\bfseries,
    boxrule=1.5pt,
    width=\linewidth,
    before skip=1em,
    after skip=1em,
    breakable,
    left=1mm, right=1mm, boxsep=1mm
]
\begingroup
\footnotesize\ttfamily
\begin{Verbatim}[
  breaklines=true,
  breakanywhere=true,
  breaksymbolleft={},
  breaksymbolright={}
]
You are an analyst that converts concrete Hotpot-style QA failures into reusable abstract error patterns for synthetic data generation.

You will receive ONE item in JSON describing (some fields may be empty):
- question, gold_answer, model_answer, type, level
- model_explanation and (possibly truncated) model_raw
- model_trace: a reconstructed 2–4 step evidence trace that matches the model_answer
- gold_supporting_facts
- missing_gold_evidence_text: gold-required evidence sentences the model did NOT use (may be empty)
- extra_pred_evidence_text: evidence sentences the model used but are NOT gold (may be empty)
- difficulty_tags and instance_error_brief
- pattern_anchor_evidence_text: key sentence(s) that directly illustrate the pitfall (may be empty)
- contexts: Wikipedia-like passages with titles and sentences (may be partial)

Your task:
Produce ONE JSON object with the following keys ONLY:
{
  "error_type": "<one short snake_case label>",
  "abstract_error_name": "<short name>",
  "abstract_error_description": "<3-6 sentences>",
  "abstract_error_template": "<a structured template describing roles like Entity A / Entity B / Doc 1 / Doc 2 and what gets confused>",
  "transfer_guidance": "<5-10 bullet-like sentences (still plain text) describing how to recreate this pitfall in new fictional domains>"
}

How to analyze (CRITICAL):
1) Start from EVIDENCE SIGNALS in this priority order:
   (a) Evidence diff (preferred):
       - Use missing_gold_evidence_text vs extra_pred_evidence_text to identify what the model ignored vs what it anchored on.
   (b) If evidence diff is empty or insufficient:
       - Use pattern_anchor_evidence_text to locate the single most “tempting” sentence that caused the mistake.
       - If pattern_anchor_evidence_text is also empty, look into contexts + gold_supporting_facts + model_trace to infer the likely confusion point (but do NOT invent evidence not present).
2) Use model_explanation/model_raw/model_trace ONLY as supporting signals:
   - Align them with the evidence signals to infer the most plausible failure mechanism (e.g., constraint dropped, ambiguity unresolved, wrong comparison target, distractor anchoring, entity linking error, multi-value-single-choice ambiguity, type-instance confusion, overstrict unknown).
3) Keep the pattern reusable and domain-agnostic:
   - Do NOT depend on the original entity names or domain.
   - Describe roles and relations abstractly (Entity A/B, Work 1/Work 2, Attribute X, List of candidates, etc.).
4) Make the template actionable for synthesis:
   - Explicitly specify which passages/sentences are TRUE supporting vs distractors.
   - Specify what a shallow model will likely do wrong and why it is tempting.
   - If the pitfall is “question expects a single value but evidence lists multiple plausible values”, spell out the list-like structure and why selecting the wrong member is tempting.
5) Be concise and concrete. Do NOT quote long text.

Output constraints (STRICT):
- Output ONLY the JSON object, no markdown, no extra text.
- All values must be plain strings (no nested JSON).
- error_type must be short and general (e.g., "distractor_anchoring", "entity_linking_confusion", "constraint_drop", "ambiguous_title_match", "multi_value_selection", "overstrict_unknown", "partial_answer").
- Do NOT mention field names like “missing_gold_evidence_text”, “contexts”, or “pattern_anchor_evidence_text” in the output.
- Before output, do a final check that ALL 5 required keys are present and non-empty strings.
\end{Verbatim}
\endgroup
\end{tcolorbox}

\begin{tcolorbox}[
    colback=gray!10,
    colframe=darkgray,
    title=Prompt for KG K\_stress (for generating error reports),
    coltitle=white,
    fonttitle=\bfseries,
    boxrule=1.5pt,
    width=\linewidth,
    before skip=1em,
    after skip=1em,
    breakable,
    left=1mm, right=1mm, boxsep=1mm
]
\begingroup
\footnotesize\ttfamily
\begin{Verbatim}[
  breaklines=true,
  breakanywhere=true,
  breaksymbolleft={},
  breaksymbolright={}
]
Based on the Gold answer and The steps of Gold answer and Model output and The steps of Model output, analyze which facts (including entities or relations) are missing from the provided input in the gold answer reasoning chain steps, causing the gold answer to be unattainable(model output is different from the gold answer). Carefully analyze each step to identify the root cause of the model’s incorrect response. Match the identified error to one of Patterns 1–3 and generate a corresponding error report.

The final output must strictly follow the JSON format:
{
    "error_type":[],
    "root_cause":[],
    "error_details":[],
    "missing_knowledge":[],
}

Predefined Error Patterns:

Pattern 1:
Name (Subtype): Missing Entity Information
Description:The provided knowledge graph lacks one or more required entities that are essential for deriving the Gold Answer.
Although the involved entities exist in the real world and are required by the gold_answer_reasoning, they are absent from the KG, preventing the model from initiating or completing the reasoning chain.
Example:
KG:1. (CompanyA, acquired, CompanyB)
Question:Who founded CompanyB?
Issue:The entity Founder of CompanyB exists in reality but is not present in the KG.

Pattern 2:
Name (Subtype): Missing Attribute Information
Description:The knowledge graph contains the relevant entities, but lacks required attribute or property information necessary to answer the question.
The Gold Answer reasoning depends on attributes (e.g., time, role, status, numerical values) that are real-world facts but are not encoded as triples in the KG.
Example:
KG:1. (PersonA, works_at, CompanyB)
Question:What is PersonA’s job title at CompanyB?
Issue:The attribute (PersonA, job_title, ?) is missing, though such information exists in reality.

Pattern 3:
Name (Subtype): Missing Inter-Triple Relation 
Description:All required entities and individual triples are present in the knowledge graph, but the KG lacks explicit relational links or paths between triples that are necessary to connect them into a complete reasoning chain.
The missing knowledge lies in the structural or relational connections between triples, rather than in entities or attributes themselves.
Example:
KG: 1. (CompanyA, acquired, CompanyB) 2. (PersonC, founder_of, CompanyB)
Question:Which company does the founder of CompanyB belong to?
Issue:The KG lacks a relation such as (PersonC, employed_by, CompanyA), which is a real-world-valid connection needed to link the triples.

Requirements:
- "error_type": The exact English name of the selected error pattern from Pattern 1-3 (e.g., "Missing Inter-Triple Relation").
- "root_cause": A concise noun phrase that abstracts the underlying failure mechanism in KG-QA. It must be abstract, reusable, and taxonomy-oriented. Do not use verbs, complete sentences, or instance-specific references.
- "error_details": A detailed explanation of the specific reason for the model's incorrect output, derived by comparing the reasoning chains (steps) of the Gold Answer and the Model Output, and interpreting them in conjunction with the input. The explanation should be precise down to the level of entities or relations used in the solving of problems.
- "missing_knowledge": Based on the Gold answer and The steps of Gold answer and Model output and The steps of Model output, analyze which facts (including entities or relations) are missing from the provided input in the gold answer reasoning chain steps, causing the gold answer to be unattainable (model output is different from the gold answer). The missing knowledge must be factually correct. And the missing knowledge does not require all possible alternatives to be provided. Instead, it is sufficient to identify one consistent set of facts that is missing from the input knowledge graph and that is necessary and sufficient to complete the gold answer reasoning chain.

\end{Verbatim}
\endgroup
\end{tcolorbox}

\begin{tcolorbox}[
    colback=gray!10,
    colframe=darkgray,
    title=Prompt for KG R\_stress (for generating error reports),
    coltitle=white,
    fonttitle=\bfseries,
    boxrule=1.5pt,
    width=\linewidth,
    before skip=1em,
    after skip=1em,
    breakable,
    left=1mm, right=1mm, boxsep=1mm
]
\begingroup
\footnotesize\ttfamily
\begin{Verbatim}[
  breaklines=true,
  breakanywhere=true,
  breaksymbolleft={},
  breaksymbolright={}
]
Since the gold answer and the model output differ, this indicates that the model's output is incorrect. Please conduct a detailed analysis to identify the cause of the error. Based on the reasoning chains ("steps") of the Gold answer and the Model output, carefully analyze each step to identify the root cause of the model's incorrect response. Match the identified error to one of Patterns 1-7 and generate a corresponding error report. Additionally, since new examples will be generated afterward, the new examples must inherit the same error pattern. Therefore, we need to analyze the way how to transfer. The Paramemeters and Conditions in the pattern can be referred into thinking.

The final output must strictly follow the JSON format:
{
    "error_type":[],
    "root_cause":[],
    "error_details":[],
    "transfer_conditions":[],
}

Predefined Error Patterns
Pattern 1:
Name (Type): Partial Entity Recognition Error
Description: All required facts exist in the provided KG and all the implied logical constraints are correct and satisfied, but the model returns only a subset of the correct result set due to incomplete traversal, early stopping, top-k bias, or failure to aggregate all bindings.
Parameters: theta = (num_entities_expected, num_entities_retrieved)
- num_entities_expected: total number of relevant entities or facts needed for the correct answer
- num_entities_retrieved: number of entities or facts actually retrieved by the model
Trigger Conditions:
- The correct answer is a set/list (multiple entities/values).
- The question requires aggregation or collection of multiple facts/entities.
For example:
1. (TeamX, award_year, "2026")
2. (TeamX, award_year, "2027")
3. (TeamX, award_year, "2030")
Question: "In which years did PersonaA win awards?", Gold answer: 2026, 2027 and 2030, model may only output 2026 and 2027

Pattern 2:
Name (Type): Multi-hop composition failure
Description: A required multi-step reasoning chain exists, but the model breaks the chain by skipping a hop, inserting an extra hop, mis-ordering hops, or binding an intermediate variable incorrectly. The result is an invalid or incorrect derivation.
Parameters: theta = (L, hop_error_pos, d_distractor)
Trigger Conditions:
- Existence of a main reasoning path of length L (typically 2-3) between source and target entities.
- Presence of >= d_distractor semantically irrelevant or misleading alternative paths of similar length.
- The model's produced path length/order differs or intermediate binding mismatches.
For example:
1. (DrLee, affiliated_with, InstA)
2. (InstA, headquartered_in, MetroZ)
Distractors: (DrLee, collaborated_with, ProfX), (ProfX, headquartered_in, OtherCity)
Question: "Which city is DrLee's affiliated institution located in?", Gold answer: MetroZ, may fail due to distraction or incomplete multi-hop reasoning.

Pattern 3:
Name (Type): Path confusion
Description: Multiple paths connect source to target; the model selects a shorter/high-salience path that is semantically misaligned with the question, instead of the gold answer semantic reasoning path.
Parameters: theta = (L_true, L_spurious, w_conf)
Trigger Conditions:
- A true semantic path exists with L_true > L_spurious.
- The shorter path relies on weak evidence or mismatched relations.
For example:
1. (MayorK, parent_of, CouncilorM)
2. (CouncilorM, employed_by, CityHall)
Short spurious path: (MayorK, met_with, CityHallRep)
Question: "Who has a family member employed by CityHall?" Gold answer: MayorK (via path 1+2), Model may incorrectly choose CityHallRep due to shortest-path bias.

Pattern 4:
Name (Type): Relation directionality confusion
Description: The model treats a directed relation as its inverse (or uses inverse where forward is needed), flipping head/tail semantics and producing reversed traversal or wrong binding.
Parameters: theta = (hasInverse, p_inverse_distractor)
Trigger Conditions:
- Both a relation r and its inverse r^-1 (or semantically similar directional relations) appear in the local subgraph.
For example:
1. (RiverX, flows_through, RegionY)
2. (RegionY, contains, CityZ)
3. (CityZ, near_river, RiverX) (weak semantics)
Question: "Which river flows through RegionY?", model may confuse direction and answer incorrectly.

Pattern 5:
Name (Type): Temporal constraint Error
Description: The model misinterprets temporal expressions (before/after/during, inclusive boundaries), or parses/normalizes date values incorrectly (2010s -> 2010). This type of mistake is only caused by the improper temporal literal expressions not referring to the relation.
Parameters: theta = (time_ops, boundary_inclusive, time_property_confusables)
Trigger Conditions:
- Question contains temporal filtering or comparison.
For example: "after 2010" treated as ">=2010" vs ">2010".

Pattern 6:
Name (Type): Constraint coverage Error
Description: This Error should be applied only when all prior reasoning steps are otherwise correct logically with no other error patterns, and the error arises solely from missing required constraints or introducing unjustified constraints in the prediction output. If the output violates the constraints explicitly stated in the question (i.e., the constraints are recognized but incorrectly satisfied), rather than omitting required constraints or adding extra ones, the error should be classified as Type constraint violation, not this error type.
Parameters: theta = (n_constraints_gold, n_constraints_pred, constraint_drop_rate)
Trigger Conditions:
- Question contains multiple clauses/filters (time, location, role, type, etc.).
- Model output ignores one clause or adds an unrelated restriction.
For example:
Question: "Which films directed by X in 2010?"
Model returns all films directed by X (misses year), or filters to 2011 (wrong extra constraint).

Pattern 7:
Name (Type): Type constraint violation
Description: The model violates domain/range constraints or fails to enforce an expected answer type, returning entities of the wrong type (person vs location, movie vs book). Only contains the situation of misuse or mismatch of type.
Parameters: theta = (type_confusables, p_type_signal)
Trigger Conditions:
- Attribute conflicts among the relevant entities
- Candidate entities have weak or conflicting type signals.
For example:
1. (EntityA, has_title, "Shoreline")
2. (EntityB, born_in, CityC)
Context ambiguously leads the model to treat EntityA as a person instead of a work.
Question: "Who created Shoreline?", Gold answer: Artist, model may treat EntityA incorrectly as a person to output wrong answer

Requirements:
- "error_type": The exact English name of the selected error pattern from Pattern 1-7 (e.g., "Multi-hop Composition Failure").
- "root_cause": A concise noun phrase that abstracts the underlying failure mechanism in KG-QA. It must be abstract, reusable, and taxonomy-oriented. Do not use verbs, complete sentences, or instance-specific references.
- "error_details": A detailed explanation of the specific reason for the model's incorrect output, derived by comparing the reasoning chains (steps) of the Gold Answer and the Model Output, and interpreting them in conjunction with the input. The explanation should be precise down to the level of entities or relations used in the solving of problems.
- "transfer_conditions": Additionally, since new examples will be generated afterward, the new examples must inherit the same error pattern. Therefore, we need to analyze the way how to transfer. The Paramemeters and Conditions in the pattern can be referred into thinking. It should contain three parts:
  1. Summarize the error in an abstract and generalizable form (use generalized, placeholder entities (e.g., Person_A, University_X, City_Y))
  2. Explain how this error pattern can be transferred and what's the difficulty
  3. Specify what characteristics or elements must be present in the newly generated context and question to ensure the error can be reproduced.
The analysis of transfer should be included in "transfer_conditions".
\end{Verbatim}
\endgroup
\end{tcolorbox}

\begin{tcolorbox}[
    colback=gray!10,
    colframe=darkgray,
    title=Prompt for Table R\_stress (for generating error reports),
    coltitle=white,
    fonttitle=\bfseries,
    boxrule=1.5pt,
    width=\linewidth,
    before skip=1em,
    after skip=1em,
    breakable,
    left=1mm, right=1mm, boxsep=1mm
]
\begingroup
\footnotesize\ttfamily
\begin{Verbatim}[
  breaklines=true,
  breakanywhere=true,
  breaksymbolleft={},
  breaksymbolright={}
]
You are an analyst that converts concrete WTQ-style TABLE QA failures into reusable, synthesis-ready REASONING error patterns for synthetic data generation.

Scope:
- This stage is REASONING ERRORS ONLY (not missing external knowledge).
- Assume the table provided is the evidence context for the question (WTQ setting). Do NOT reject because an entity name is not in a column; the table itself is the context.
- If the model answered "unknown" / refused / claimed insufficiency when the answer is actually computable from the table, classify it as a reasoning error (overstrict_unknown / evidence_scope_misinterpretation).

You will receive ONE instance containing:
- id, question, gold_answer, model_answer
- table as JSON (header + rows) or table_md
- model_explanation (optional)

Your task:
Return ONE JSON object with EXACTLY these keys and no others:
{
  "case_id": "<string>",
  "reasoning_family": "<enum>",
  "required_ops": "<comma-separated ops from vocab as a STRING>",
  "bottleneck_step": "<enum>",
  "trigger": "<VERY detailed, synthesis-actionable trigger>",
  "error_signature": "<short description of the wrong behavior/signature>",
  "evidence_spec": "<plain-text specification of needed columns + key cells/row selectors>",
  "ambiguity": "<'false' or 'true: ...' as a STRING>",
  "transfer_guidance": "<5-10 bullet-like lines in ONE string; each line starts with '- ' and lines separated by \\n>"
}

Enums:
reasoning_family $\in$
- lookup_single_cell
- filter_then_lookup
- filter_then_aggregate
- aggregate_then_compare
- normalize_then_compare
- compute_difference_or_ratio
- argmax_argmin
- temporal_ordering
- string_parse_then_reason
- multi_hop_composition

bottleneck_step $\in$
- unit_normalization
- format_parsing
- aggregation
- comparison_argmin_argmax
- arithmetic
- filtering_logic
- temporal_ordering
- multi_step_composition
- string_parse_then_reason
- other_reasoning

Ops vocab (pick all that apply, output as a comma-separated STRING):
- select_rows
- select_columns
- filter_rows
- lookup_cell
- normalize_unit
- parse_number
- aggregate_sum
- aggregate_avg
- aggregate_count
- compare_argmin
- compare_argmax
- sort_by_time
- compute_difference
- compute_ratio
- compute_percent

CRITICAL: trigger must be VERY DETAILED and actionable for synthesis.
Write trigger as ONE string with these 4 labeled sections IN ORDER (use labels exactly):
1) Surface pattern:
2) Reasoning requirement:
3) Failure signature:
4) Anti-ambiguity constraints:

- Surface pattern must describe concrete, checkable table/question cues (e.g., a year appears in a Date column; counting is required; entity only appears in the question and not as a column).
- Reasoning requirement must describe the minimal operation sequence.
- Failure signature must describe the typical wrong behavior (e.g., unjustified "unknown" due to misinterpreting table scope).
- Anti-ambiguity constraints must state how to guarantee a unique answer (e.g., define the year filter precisely; ensure all rows are within the relevant context; avoid mixed-year rows unless intended).

evidence_spec must be plain text and include:
- needed_columns: ...
- key_rows: ...
- how_to_compute: ...

ambiguity must be a STRING:
- "false" if unique answer
- "true: <reason>" if multiple valid answers

transfer_guidance must:
- Preserve the SAME reasoning_family + ops pattern.
- Instruct how to synthesize new WTQ-style questions/tables that recreate the pitfall.
- Include invariants, distractors/traps, and checks to avoid ambiguity.
- Be domain-agnostic (no original entity names).

Output constraints:
- Output ONLY the JSON object (no markdown).
- All values must be plain strings (no arrays/objects).
- Ensure all required keys are present and non-empty strings.

###USER_TEMPLATE
Here is the instance JSON:
{input_ison}
\end{Verbatim}
\endgroup
\end{tcolorbox}

\subsection{Data Synthesis}

\begin{tcolorbox}[
    colback=gray!10,
    colframe=darkgray,
    title=Prompt for Text K\_stress (for synthesizing data),
    coltitle=white,
    fonttitle=\bfseries,
    boxrule=1.5pt,
    width=\linewidth,
    before skip=1em,
    after skip=1em,
    breakable,
    left=1mm, right=1mm, boxsep=1mm
]
\begingroup
\footnotesize\ttfamily
\begin{Verbatim}[
  breaklines=true,
  breakanywhere=true,
  breaksymbolleft={},
  breaksymbolright={}
]
You are a dataset augmentation assistant for question answering.

You will be given a single HotpotQA-style example containing:
- ORIGINAL QUESTION
- ORIGINAL GOLD ANSWER (for synthesis only; DO NOT reveal it in any new question)
- CONTEXTS: a list of Wikipedia contexts, each has fields like title, paragraph/sentences, and is_supporting true/false
- In the user message, you will also be told:
  * how many items to output in THIS call
  * a target quota/distribution and how many have already been accepted for this seed

Your job:
Generate EXPANDED QA items that add ONE extra reasoning hop beyond the original question, with strict correctness.

The new items should be designed to probe external-knowledge gaps:
- The new question must indirectly depend on the ORIGINAL GOLD ANSWER (the hidden first hop).
- The new gold answer must be strictly correct.
- A model that does NOT know the hidden first-hop fact is likely to fail when answering ONLY the new question (especially in no-context evaluation).

You MUST follow ALL rules below.

=======================
HARD OUTPUT CONSTRAINTS
=======================
1) Output MUST be exactly ONE JSON object and nothing else (no markdown, no code fences).
2) Output schema must be exactly:
   {"items":[{"new_question":"...","recipe":"...","new_gold_answer":"...","clue":"..."}, ...]}
3) Output exactly the number of items requested in the user message for this call (no more, no less).
4) Do NOT output any extra keys beyond: new_question, recipe, new_gold_answer, clue.

=====================
ANTI-LEAK CONSTRAINTS
=====================
5) NEVER include the ORIGINAL GOLD ANSWER as a substring in any new_question (case-insensitive).
6) NEVER include the ORIGINAL GOLD ANSWER inside the clue for SAME items (SAME clue must be empty anyway).
7) For any item where recipe starts with "HOP:", new_gold_answer MUST be DIFFERENT from the original gold answer.

====================
ALLOWED RECIPE TYPES
====================
There are exactly two recipe families:

A) SAME (answer unchanged; expanded/indirect phrasing)
- recipe: "SAME"
- new_gold_answer: MUST equal the original gold answer exactly
- clue: MUST be "" (empty string)
- new_question:
  * must NOT be a trivial paraphrase
  * must add an extra reasoning layer in wording (more indirect), but the correct answer remains the same
  * must NOT introduce new factual claims beyond what is already implied by the original question

B) HOP (ENTITY_DERIVED_1HOP_WITH_EVIDENCE)
- recipe format: "HOP:<RELATION>"
- The relation MUST be single-valued, unambiguous, and explicitly supported by the provided CONTEXTS.
- You MUST use ONLY the provided CONTEXTS to justify this hop (do NOT rely on outside knowledge for correctness).
- new_question:
  * hides the first hop by referring to the original gold entity indirectly through the original question’s concept
  * asks for exactly ONE derived value (one hop away from the gold entity)
  * must NOT reveal the original gold answer
- new_gold_answer:
  * the derived 1-hop value
  * MUST differ from the original gold answer
- clue (MANDATORY for HOP):
  * MUST be an exact contiguous substring copied from the provided CONTEXTS (character-for-character)
  * MUST contain new_gold_answer verbatim
  * MUST directly support that new_gold_answer is the correct derived value
  * keep clue short (<= 180 characters)
  * do NOT paraphrase the clue; do NOT add ellipses or brackets; copy exactly

Preferred RELATION choices (use only when clearly present and single-valued in CONTEXTS):
- film_title_about_entity / work_title_about_entity
- directed_by / written_by / author_of_work
- language_of_work
- year_of_release
- located_in_country / located_in_city / located_in_state
- capital_of_country (only if explicitly stated)
Avoid multi-valued relations unless the CONTEXTS explicitly indicate a single answer.

======================
QUOTA / DISTRIBUTION (VERY IMPORTANT)
======================
The user message will state a target distribution (e.g., how many SAME vs HOP remain) and how many have already been accepted.
You MUST:
- Prefer producing items that satisfy the remaining quota.
- Avoid producing items that exceed the remaining quota.
- If you are unsure a HOP is single-valued and explicitly supported, do NOT produce that HOP; produce a SAME item instead.

======================
DIVERSITY RULES
======================
Within the items you output in this call:
- Items must be clearly different, not minor rewordings.
- Use at least 3 different surface forms among them:
  * different question structures (wh-question vs. identification vs. constraint framing)
  * different bridging/relational phrasing
  * different indirect reference styles
- If two questions differ by only 1–3 words, it is INVALID.

======================
NO INVENTION / SAFETY
======================
- Do NOT invent facts.
- For HOP items, the derived value must be supported by the copied clue from CONTEXTS.
- Do not output explanations outside the JSON.

Return exactly one JSON object matching the schema.
\end{Verbatim}
\endgroup
\end{tcolorbox}

\begin{tcolorbox}[
    colback=gray!10,
    colframe=darkgray,
    title=Prompt for Text R\_stress (for synthesizing data),
    coltitle=white,
    fonttitle=\bfseries,
    boxrule=1.5pt,
    width=\linewidth,
    before skip=1em,
    after skip=1em,
    breakable,
    left=1mm, right=1mm, boxsep=1mm
]
\begingroup
\footnotesize\ttfamily
\begin{Verbatim}[
  breaklines=true,
  breakanywhere=true,
  breaksymbolleft={},
  breaksymbolright={}
]
You are a synthetic data generator for multi-hop reading comprehension tasks similar to the HotpotQA dataset.

TOP PRIORITIES (MUST follow this order):
1) PATTERN FIDELITY (MOST IMPORTANT): Faithfully instantiate the SAME error pattern described by the input
   "abstract_error_template" and "transfer_guidance".
2) EVIDENCE ALIGNMENT + CLOSED WORLD: The gold answer must be uniquely derivable ONLY from the provided contexts,
   and the required evidence must be exactly captured by supporting_facts (title + 0-based sent_id).
3) MULTI-HOP DIFFICULTY: The question must require >=2 reasoning steps across at least TWO different context titles,
   and must strongly tempt the target error pattern.
4) FICTIONAL PROPER NAMES (NO PLACEHOLDERS REQUIRED): Use invented fictional proper names consistently.
   Absolutely NO real-world entities of any kind.
5) STRICT JSON OUTPUT: Output one valid JSON object matching the required schema, with no extra text.

INPUT:
You will receive ONE error_report JSON that includes (among other fields):
- abstract_error_template
- transfer_guidance
- error_type_canon (may exist)
You MUST treat abstract_error_template and transfer_guidance as authoritative.

YOUR TASK:
Generate ONE NEW, fully synthetic Hotpot-style QA example in a DIFFERENT fictional domain than the input.
Do NOT reuse any concrete names, places, titles, or surface strings from the input item.

HARD CONSTRAINT: NO REAL-WORLD ENTITIES (ABSOLUTELY REQUIRED)
- Do NOT use any real people/countries/cities/companies/brands/universities/awards/media titles, etc.
- Do NOT use obvious real-world strings like "United States", "China", "London", "Harvard", "Apple", "Nobel", "Marvel", "Disney", etc.
- If you are unsure whether a name might be real, invent a new one instead.
- This applies to question, gold_answer, all context titles, all sentences, and pattern_application.

QUALITY GUARDRAILS (keep output short but high-quality):
- The 2-hop reasoning must be REAL:
  Hop1 identifies an intermediate entity/relation from one supporting context,
  Hop2 uses a DIFFERENT supporting context to obtain/derive the final answer.
- At least ONE distractor passage must contain a tempting near-match that triggers the target error:
  (same relation type, similar attribute, more salient phrasing, or type/alias confusion).
- Avoid vacuous worldbuilding: every supporting sentence must be necessary; distractors must be plausible but not required.
- Include at least one disambiguating clue that only careful readers will use (qualifier, alias note, scope boundary, “not to be confused with” style).

DIVERSITY REQUIREMENTS (VERY IMPORTANT; reduce repetitive templates):
You MUST vary these dimensions across generations:
A) Question operator: choose ONE operator type that fits the pattern but AVOID overusing yes/no “both before YEAR” style.
   Pick from: {who/which person, where, which organization, what role/title, which year/date, which item, how many/number, which of two, comparison}.
B) Relationship type between hops:
   Pick ONE: {authored-by, founded-by, located-in, member-of, played-by, directed-by, invented-by, chaired-by, issued-by, recorded-in, won-by, named-after}.
C) Answer type:
   Choose ONE: {person name, organization name, place name, short role/title phrase, single number, year/date}.
D) Distractor style:
   Choose ONE main distractor mechanism aligned with the target pattern:
   - salient wrong passage
   - near-duplicate entity/title collision
   - alias/surface-form mismatch
   - multi-value enumeration in one sentence
   - misleading type word in the question
   - reference/pronoun ambiguity
   - constraint/qualifier hidden in a less-salient sentence
You MUST NOT repeat the same operator+relationship+answer-type combination as the input seed’s surface structure.

SIZE / STRUCTURE:
- contexts: exactly 6 to 9 passages.
- each passage: 2 to 4 sentences.
- at least 2 passages must be is_supporting=true.
- supporting_facts: 2 to 5 items, each referencing only is_supporting=true titles.
- gold_answer must be short and unique.

OUTPUT JSON SCHEMA (STRICT):
Output exactly ONE JSON object with the following top-level keys ONLY:

{
  "question": "<English question>",
  "gold_answer": "<short answer>",
  "contexts": [
    {
      "title": "<Wikipedia-style fictional title>",
      "sentences": ["sentence 0.", "sentence 1.", "..."],
      "paragraph": "<exactly the sentences joined by single spaces>",
      "is_supporting": true or false
    }
  ],
  "supporting_facts": [
    { "title": "<supporting context title>", "sent_id": <0-based integer> }
  ],
  "pattern_application": "<MANDATORY bullet-like explanation>"
}

FIELD RULES:
- question:
  - Must require >=2 hops across >=2 different titles.
  - Must tempt the SAME failure mode as the input pattern.
  - Must not mention HotpotQA, data generation, “pattern”, “error”, or anything meta.

- gold_answer:
  - Must be uniquely correct from contexts.
  - Must be short (name, short title/role phrase, number, or year/date).

- contexts:
  - Each title must be fictional.
  - sentences must be grammatical English.
  - paragraph must exactly equal sentences joined by single spaces (no extra changes).

- supporting_facts:
  - Each must reference a title where is_supporting=true.
  - sent_id must be valid for that title’s sentences (0-based).
  - The union of these sentences must be sufficient to derive gold_answer uniquely.

PATTERN_APPLICATION (MANDATORY):
Write 6–10 bullet-like lines (each line starts with "- "):
- Map template roles to your fictional entities (Entity A/B, Doc1/Doc2, Attribute X, etc.).
- Describe Hop1 and Hop2 and cite which titles contain each hop.
- List which supporting_facts are necessary and why.
- Identify the main distractor passage(s) and the tempting cue.
- State the target mistake a shallow model would make (wrong answer form/type/constraint) and why it is tempting.
- State the key constraint/disambiguator and where it appears.
- Confirm: pattern fidelity, closed-world uniqueness, and no real-world entities.

OUTPUT CONSTRAINTS (CRITICAL):
- Output ONLY the JSON object.
- No markdown fences.
- No additional commentary.
- Keep the whole output compact; avoid long lists and avoid more than 4 sentences per passage.
\end{Verbatim}
\endgroup
\end{tcolorbox}

\begin{tcolorbox}[
    colback=gray!10,
    colframe=darkgray,
    title=Prompt for KG K\_stress (for synthesizing data),
    coltitle=white,
    fonttitle=\bfseries,
    boxrule=1.5pt,
    width=\linewidth,
    before skip=1em,
    after skip=1em,
    breakable,
    left=1mm, right=1mm, boxsep=1mm
]
\begingroup
\footnotesize\ttfamily
\begin{Verbatim}[
  breaklines=true,
  breakanywhere=true,
  breaksymbolleft={},
  breaksymbolright={}
]
You are a question generation model for knowledge graph QA.

The final output must strictly follow the JSON format:
{
    "New_example_question": [],
    "New_example_gold_answer": []
}

Inputs:
- "i": the number of question-answer pairs (New_example_question, New_example_gold_answer)
- "Input_KG": the structured knowledge triples provided as input.
- "missing_knowledge": the additional triples that are needed to answer the questions correctly.

Task:
- Using ONLY the provided Input KG triples and the missing_knowledge triples, generate i different questions.
- Each question must require reasoning that uses one or more missing_knowledge triples.
  If a question can be answered correctly using only the Input KG (without missing_knowledge), it is INVALID.
- No direct, indirect, or shortcut reasoning paths may exist in the Input KG that connect the entities needed to answer the question without missing_knowledge.
- Do NOT directly ask for or rephrase any fact contained in missing_knowledge triples.
- Each question must rely on a unique reasoning path or distinct combination of triples from the merged KG (Input KG + missing_knowledge).
- Questions must be concise and clear; do not include guiding words or hints like "By", "Consider", or similar.
- Answers must be unique, fully correct, and retrievable ONLY from the merged KG.
- Do NOT use any external knowledge beyond the merged KG.
- Do NOT include the answer in the question.
- Each question should be syntactically and semantically distinct from the others.

Output requirements:
- Output a single JSON object in the exact format below.
- Only the JSON object. Do NOT include markdown, explanations, or extra text.
- The arrays must be aligned by index, i.e., the i-th question corresponds to the i-th answer.
\end{Verbatim}
\endgroup
\end{tcolorbox}

\begin{tcolorbox}[
    colback=gray!10,
    colframe=darkgray,
    title=Prompt for KG R\_stress (for synthesizing data),
    coltitle=white,
    fonttitle=\bfseries,
    boxrule=1.5pt,
    width=\linewidth,
    before skip=1em,
    after skip=1em,
    breakable,
    left=1mm, right=1mm, boxsep=1mm
]
\begingroup
\footnotesize\ttfamily
\begin{Verbatim}[
  breaklines=true,
  breakanywhere=true,
  breaksymbolleft={},
  breaksymbolright={}
]
Referring to the error_details and Transfer conditions for this error type, generate 1 fully independent knowledge graph QA examples that inherit the error from the prior analyzed case and ouput the gold answer of this New_example, strictly following the Transfer_Conditions for that error type.
The final output must strictly follow the JSON format:
{
    "New_example":[],
    "New_example_gold_answer":[]
}

Referring to the three parts in Transfer codnitions for this error type -- (1) Abstracted error form, (2) how to transfer/difficulty, and (3) Required characteristics for new examples -- and generate 1 new examples that successfully transfers the old error pattern to the new example. The new examples must satisfy the following requirements:
You are asked to generate 1 fully independent knowledge graph QA examples that inherit the error from the prior analyzed case.

- - - - - - - - - - - - - - - - - - - - - - - - - - - - - - - - - - - - -
Requirements for the generated examples:
- - - - - - - - - - - - - - - - - - - - - - - - - - - - - - - - - - - - -
Carefully analyze the provided fields transfer_conditions, where transfer_conditions contains three subfields: Abstract Error Form, Transfer Mechanism & Difficulty, and Required Characteristics for New Context & Question.
The goal is to analyze the specific reasons why errors occurred in the original example, and then reproduce this error in a new example based on the abstract pattern, and required characteristics, using placeholders for all intermediate entities and relations.
***All intermediate entities and relations in New_example_input (including "question:" section) and New_example_gold answer MUST be the placeholders.***

1. error_details:
- Purpose: Explain exactly why the model failed, comparing Gold Answer reasoning vs model output.
- What to learn for new example: Identify which entity, relation, or combination caused the error.
- How to apply: The new KG must include these critical elements or relations so that the error can be reproduced. This guides the placement of distractors, intermediate nodes, and multi-condition dependencies.

2. transfer_conditions:
- Purpose: Specify how the error pattern can be transferred to a new context.
- Structure:
  a) Abstract Error Form:
     - Role: Summarize the logical structure of the error with placeholder entities and relations (e.g., Person_A, Country_X, Organization_Y).
     - Guidance: New examples must reproduce this logical structure exactly, including the intermediate nodes or shared-path elements that can induce early-stopping errors.
  b) Transfer Mechanism & Difficulty:
     - Role: Explain why the error occurs and what makes it hard for the model to reason correctly.
     - Guidance: New examples should maintain the same multi-hop dependencies, distractor paths, and cross-type aggregation that generate reasoning difficulty.
  c) Required Characteristics for New Context & Question:
     - Role: Define the structural and relational features the new KG and question must have to inherit the error pattern.
     - Guidance: Include multiple entity types and relation types, ensure shared early hops for distractors, enforce multi-condition constraints, and design the question such that only the full reasoning chain leads to the Gold Answer.

****New example Output format*****:
- Format exactly like the original examples:
  - Start with: "You will be given a list of knowledge graph triples. Answer the following question using the information in the triples."
  - Include a section "knowledge graph triples:" with triples separated by "|".
    - The total number of triples must be between 30 and 40; if the original triples are insufficient, additional triples may be added only if they are strongly related to the existing context (e.g., same entity types, same hierarchical relations, parallel distractor paths) and must not alter, weaken, or repair the inherited error pattern or provide information that could help the model complete correct reasoning.
    - Each triple must have exactly three parts: ENTITY RELATION VALUE
      Only three RELATION types are allowed. Ignore all others, including single entities without a relation:
      1) Attribute Relations (entity_attribute)
         - VALUE is a literal (not another entity).
         - Form: entity_attribute or entity.attribute.
         - Examples: city_population, state_governor, economy_gdp.
      2) Entity-to-Entity Relations (entity_relation_entity)
         - VALUE is another entity.
         - Form: entity_relation_entity or entity.verb.entity.
         - Examples: located_in, has_state, scientist_affiliated_with.
      3) Hierarchical Attribute Relations (entity.entity.attribute)
         - VALUE is typically a literal.
         - Form: entity.entity.attribute.
         - Examples: location.country.capital, location.country.language.
    - Use a single space between ENTITY, RELATION, and VALUE
    - Do NOT use "|" inside a triple
    - Separate triples using "|" as a delimiter
  - Include a section "question:" with a multi-hop question at the end. (no hints or information in the parentheses)
  - MUST use placeholders for all intermediate entities and relations (e.g., Country_X, Province_Y, City_Z, Org_W, Party_V), except for the main value entities that should remain real/meaningful to preserve the error pattern (e.g., French, Standard French, English).

- - - - - - - - - - - - - - - - - - - - - - - - - - - - - - - - - - - - -
Requirements for the New_example_gold_answer:
- - - - - - - - - - - - - - - - - - - - - - - - - - - - - - - - - - - - -
1) You must generate the correct answer (gold_answer) for the "New_example" and store it in "New_example_gold_answer".
2) To ensure the correctness and executability of the gold_answer, you must perform internal step-by-step reasoning when generating the answer; however, do NOT output the reasoning process.
3) You must refer to the specified error_type, details and the transfer_conditions, and explicitly avoid the described failure cases/situation when reasoning and constructing the gold_answer, so that the output represents the correct logical form for subsequent evaluation.
4) If multiple correct answers exist, separate them with commas. If doesn't exist, return "None".
Do NOT output anything other than the correct answer(s) -- this includes explanations, evidence, reasoning steps, or any additional text.

Output: Generate 1 separate KG-QA examples (with question) and its gold_answer following the above rules. Each example MUST inherit the error pattern and should be complex, with strong multi-path interference, cross-type aggregation, multi-condition constraints, and heavy noise. And the gold_answer should be the correct answer to the question in New_example.
\end{Verbatim}
\endgroup
\end{tcolorbox}

\begin{tcolorbox}[
    colback=gray!10,
    colframe=darkgray,
    title=Prompt for Table R\_stress (for synthesizing data),
    coltitle=white,
    fonttitle=\bfseries,
    boxrule=1.5pt,
    width=\linewidth,
    before skip=1em,
    after skip=1em,
    breakable,
    left=1mm, right=1mm, boxsep=1mm
]
\begingroup
\footnotesize\ttfamily
\begin{Verbatim}[
  breaklines=true,
  breakanywhere=true,
  breaksymbolleft={},
  breaksymbolright={}
]
You are a synthetic data generator for table-based QA tasks similar to WikiTableQuestions (WTQ).

TOP PRIORITIES (MUST follow this order):
1) PATTERN FIDELITY (MOST IMPORTANT):
   Faithfully instantiate the SAME reasoning error pattern described by the input fields:
   "trigger" and "transfer_guidance" (authoritative).
2) CLOSED WORLD + UNIQUE ANSWER:
   The gold answer must be uniquely derivable ONLY from the provided table.
   Do NOT require any external knowledge.
3) TRAP STRENGTH:
   The table + question must strongly tempt the target failure signature implied by the pattern
   (e.g., overstrict unknown, filter drop, wrong argmax target, unit normalization skip).
4) FICTIONAL PROPER NAMES (ABSOLUTELY REQUIRED):
   Use invented fictional proper names consistently.
   Absolutely NO real-world entities of any kind.
5) STRICT JSON OUTPUT:
   Output one valid JSON object matching the required schema, with no extra text.

INPUT:
You will receive ONE "pattern_seed" JSON object that includes:
- case_id
- reasoning_family
- required_ops
- bottleneck_step
- trigger
- transfer_guidance
(Other fields may exist. Treat trigger + transfer_guidance as authoritative.)

HARD CONSTRAINT: NO REAL-WORLD ENTITIES (ABSOLUTELY REQUIRED)
- Do NOT use any real people/countries/cities/companies/brands/universities/awards/media titles, etc.
- Do NOT use obvious real strings like "United States", "China", "London", "Harvard", "Apple", "Nobel", "Marvel", "Disney", etc.
- If you are unsure whether a name might be real, invent a new one instead.
- This applies to ALL strings in the output: question, header, cells, row values, and pattern_application.

TABLE QA REQUIREMENTS:
- The output must look like a real WTQ-style table with header + rows.
- Use 6 to 14 rows (avoid tiny tables).
- Use 4 to 9 columns.
- Include at least one distractor mechanism aligned with the pattern (e.g., near-match rows, similar roles, mixed formats, misleading ordering).
- Ensure the question is solvable with the intended operations and produces a UNIQUE gold_answer.
- Ensure the question is answerable from the table ALONE (closed-world).

DIVERSITY REQUIREMENTS (VERY IMPORTANT):
You MUST vary these dimensions across generations:
A) Question operator: choose ONE from:
   {how many/number, which item/name, what value, which of two, comparison (highest/lowest), earliest/latest}
B) Domain style: choose ONE fictional domain each time:
   {fictional sports season log, fictional festival schedule, fictional lab experiment results, fictional library inventory,
    fictional shipping ledger, fictional academy course roster, fictional game tournament standings, fictional vehicle service log}
C) Trap style (choose ONE main trap aligned with the pattern):
   - scope_misinterpretation (table is the context, but entity name not present as a column)
   - filter_drop (model ignores a qualifier like year/category)
   - wrong_target_column (compares wrong numeric column)
   - argmin_argmax_flip
   - unit_normalization_skip (mixed units or per-period fields)
   - string_number_parse_error (commas, currency, fractions)
   - temporal_order_shortcut (unsorted time column)
D) Answer type: choose ONE:
   {single number, short name, short label, short date/year}

STRICT JSON OUTPUT (CRITICAL):
- Output EXACTLY ONE JSON object.
- Output ONLY the JSON object (no markdown, no code fences, no commentary, no extra keys).
- If you cannot satisfy ALL required fields and constraints, you MUST still output a complete JSON object; do not output partial JSON.
- NEVER output empty strings for any field.

HARD FAIL CONDITIONS (DO NOT VIOLATE):
- Do NOT output any top-level keys other than the required 5 keys.
- Do NOT output any empty string fields.
- pattern_application must NOT be "" / "N/A" / "-" / "unknown" / "none".
- pattern_application must be at least 120 characters total AND contain 6–10 lines.
- supporting_cells must be 2–8 items with valid 0-based indices in range.

OUTPUT JSON SCHEMA (STRICT):
Output exactly ONE JSON object with the following top-level keys ONLY:

{
  "table": {
    "header": ["<col1>", "<col2>", "..."],
    "rows": [
      ["<r1c1>", "<r1c2>", "..."],
      ["<r2c1>", "<r2c2>", "..."]
    ]
  },
  "question": "<English question>",
  "gold_answer": "<string answer>",
  "supporting_cells": [
    { "row": <0-based row index>, "col": <0-based col index>, "why": "<short reason>" }
  ],
  "pattern_application": "<6–10 lines; each line starts with '- ' and lines separated by \\n>"
}

FIELD RULES:
- table:
  - header must be a list of 4–9 strings.
  - rows must be a list of 6–14 rows.
  - each row must have exactly the same number of cells as header.
  - all cells must be strings (numbers should be represented as strings too).
  - include at least one distractor aligned with the trap (near matches, misleading ordering, mixed formats, etc.).

- question:
  - Must be answerable from the table alone (closed world).
  - Must instantiate the SAME pattern as described by trigger + transfer_guidance.
  - Must not mention data generation, “pattern”, “error”, “trap”, “model”, or anything meta.
  - Must be clear and unambiguous; avoid multiple valid answers.

- gold_answer:
  - Must be uniquely correct from the table.
  - Must be short and correspond to EXACTLY one correct result (no lists, no multiple answers).
  - Must be consistent with the question and table.

- supporting_cells:
  - Provide 2 to 8 cells sufficient to derive the gold_answer uniquely.
  - row/col are 0-based indices into table["rows"] and table["header"].
  - why must be a short phrase (e.g., "year filter", "value to compare", "tie-break evidence").
  - supporting_cells must not be out of bounds.

- pattern_application (MANDATORY, DO NOT LEAVE EMPTY):
  - MUST be a non-empty string (>= 120 characters).
  - MUST contain 6–10 lines separated by "\n".
  - Each line MUST start with "- " (dash + space).
  - MUST mention at least one column name exactly as in table["header"].
  - MUST mention at least one 0-based row index number.
  - MUST be specific to THIS instance (no generic filler).
  - MUST confirm no real-world entities.

PATTERN_APPLICATION CONTENT TEMPLATE (FOLLOW THIS TEMPLATE; FILL IN SPECIFICS):
Write 6–10 lines, each starting with "- ", and separated by "\n".
Every line must be concrete and reference this instance (mention at least one header name and one row index somewhere).

Use this template:
- Reasoning family: <reasoning_family>; Required ops: <required_ops>.
- Chosen settings: operator=<A>, domain=<B>, trap=<C>, answer_type=<D>.
- Trigger instantiation: <how the trigger is realized in the question/table>.
- Transfer guidance used: <one sentence paraphrase of transfer_guidance applied here>.
- Minimal solution steps: <2–3 steps to get gold_answer from the table>.
- Main trap: <trap style> via <distractor cue> (cite a header name and a row index).
- Distractors: <which near-match rows/cells mislead> (cite row indices).
- Uniqueness: <why only one answer is correct> (mention tie-break if relevant; cite header/row).
- Fictional check: confirm all proper names are invented and closed-world holds.

FINAL CHECKLIST (SILENTLY VERIFY BEFORE OUTPUT):
- Exactly 5 top-level keys and no extras.
- table has 6–14 rows and 4–9 columns; all strings; consistent row length.
- question answerable from table only; unique answer.
- gold_answer is short and uniquely correct.
- supporting_cells: 2–8 items; indices valid; why non-empty.
- pattern_application: 6–10 lines, each starts "- ", contains at least one header name and one 0-based row index, and is NOT empty.
- No real-world entities anywhere.

OUTPUT CONSTRAINTS (CRITICAL):
- Output ONLY the JSON object.
- No markdown fences.
- No additional commentary.
- JSON strings may include "\n" line breaks (especially pattern_application); do not omit required content for compactness.
- Keep the whole output realistic and not overly verbose.
\end{Verbatim}
\endgroup
\end{tcolorbox}

\section{Instances in \textsc{StressEval}}
\label{sec:instance}

In this section, we provide representative instances of \textsc{Dynamic-OneEval} for each knowledge Type.

\subsection{Text Reasoning Instances}

\begin{tcolorbox}[
    colback=blue!4,
    colframe=blue!55!black,
    colbacktitle=blue!55!black,
    title=Synthetic Data (Text K\_stress) — Sample 1,
    coltitle=white,
    fonttitle=\bfseries,
    boxrule=1.5pt,
    width=\linewidth,
    before skip=1em,
    after skip=1em,
    breakable,
    left=1mm, right=1mm, boxsep=1mm
]
\begingroup
\footnotesize\ttfamily
\begin{Verbatim}[
  breaklines=true,
  breakanywhere=true,
  breaksymbolleft={},
  breaksymbolright={}
]
==============================
Original Sample (Original QAS)
==============================

S (Context):
Carl Barks (March 27, 1901 -- August 25, 2000) was an American cartoonist, author, and painter.
He is best known for his comics about Donald Duck and as the creator of Scrooge McDuck.
He worked anonymously until late in his career; fans dubbed him The Duck Man and The Good Duck Artist.

Donald Duck is a cartoon character created in 1934 at Walt Disney Productions.
Donald is an anthropomorphic white duck with a yellow-orange bill, legs, and feet.
He typically wears a sailor shirt and cap with a bow tie.
Donald is most famous for his semi-intelligible speech and his mischievous and temperamental personality.

Q (Question):
Carl Barks is best known for his comics about the cartoon character created in what year?

A (Answer):
1934


==========================
New Sample (Synthetic QAS)
==========================

S (Context):
Scrooge McDuck is a fictional character created in 1947 by Carl Barks during his time as a work-for-hire for The Walt Disney Company.
Scrooge is an elderly Scottish anthropomorphic Pekin duck with a yellow-orange bill, legs, and feet.
He typically wears a red or blue frock coat, top hat, pince-nez glasses, and spats.
He is portrayed in animations as speaking with a Scottish accent.

Music Inspired by the Life and Times of Scrooge is the first solo album by Finnish songwriter and keyboardist Tuomas Holopainen, best known for his work in the symphonic metal band Nightwish.
It was based on cartoonist Don Rosa's The Life and Times of Scrooge McDuck, a graphic novel which featured the Carl Barks Disney comics character of the same name.
Rosa contributed the cover artwork.
The first single, A Lifetime of Adventure was released on February 5, 2014 along with a music video directed by Ville Lipiäinen.

Q (Question):
From the album whose title refers to the Life and Times of the duck created by Carl Barks, who directed the music video released alongside the first single?

A (Answer):
Ville Lipiäinen


================
Difficulty Card
================

Error Type:
Closed-book knowledge gap + multi-hop bridge reasoning (cross-domain linking)

Why it is difficult:
- Indirect entity reference: the album is referred to descriptively ("the Life and Times of ..."), requiring mapping to the exact album title.
- Multi-hop bridging across entities and artifacts:
  Carl Barks -> Scrooge McDuck -> The Life and Times of Scrooge McDuck -> album -> first single -> music video director.
- Cross-domain shift: from comics/character knowledge to music metadata (single + music video director).
- Low-salience proper name: the director name is hard to guess and easy to hallucinate without explicit evidence.

How to convert / What changed:
Conversion strategy:
- Add one hop by introducing an intermediate artifact that points to an album, then query a downstream attribute explicitly stated in context.

Key modifications:
- S: add supporting context about the album and include the explicit clue sentence naming the director of the music video released alongside the first single.
- Q: rewrite the question to refer to the album indirectly and ask for the director of the first single's music video.
- A: ensure the answer is uniquely recoverable from the explicit statement in the added context.

What to check (evaluation focus):
- The model must identify the target album from the paraphrased description and extract the director from the first-single music-video sentence.
- Pitfalls: confusing Donald Duck vs Scrooge McDuck; answering with Tuomas Holopainen or Don Rosa instead of the director; hallucinating a director without context.


\end{Verbatim}
\endgroup
\end{tcolorbox}
\begin{tcolorbox}[
    colback=blue!4,
    colframe=blue!55!black,
    colbacktitle=blue!55!black,
    title=Synthetic Data (Text K\_stress) — Sample 2,
    coltitle=white,
    fonttitle=\bfseries,
    boxrule=1.5pt,
    width=\linewidth,
    before skip=1em,
    after skip=1em,
    breakable,
    left=1mm, right=1mm, boxsep=1mm
]
\begingroup
\footnotesize\ttfamily
\begin{Verbatim}[
  breaklines=true,
  breakanywhere=true,
  breaksymbolleft={},
  breaksymbolright={}
]
==============================
Original Sample (Original QAS)
==============================

S (Context):
Filming "The Trial" is an unfinished making-of film by Orson Welles, made in 1981, which focuses on the production of his 1962 film "The Trial".

The Trial (1962) is a film directed by Orson Welles, who also wrote the screenplay based on the novel of the same name by Franz Kafka.
Filmed in Europe, Welles stated immediately after completing the film: ""The Trial" is the best film I have ever made"".

Q (Question):
Orson Welles made Filming the Trial, a making-of film of the production of The Trial, which was originally filmed where?

A (Answer):
Europe


==========================
New Sample (Synthetic QAS)
==========================

S (Context):
Filming "The Trial" is an unfinished making-of film by Orson Welles, made in 1981, which focuses on the production of his 1962 film "The Trial".

The Trial (1962) is a film directed by Orson Welles, who also wrote the screenplay based on the novel of the same name by Franz Kafka.

Q (Question):
The unfinished 1981 making-of film by Orson Welles centers on the production of his 1962 movie adapted from a novel. Who wrote the novel that the 1962 film is based on?

A (Answer):
Franz Kafka


================
Difficulty Card
================

Error Type:
Closed-book knowledge gap + multi-hop bridge reasoning (making-of film -> target film -> source novel)

Why it is difficult:
- Indirect reference: the question does not directly name the 1962 film, requiring identification via the description ("1981 making-of film" -> "his 1962 movie").
- Multi-hop bridging:
  Filming "The Trial" (1981 making-of) -> The Trial (1962 film) -> based on a novel -> novel author.
- Low-salience factual detail: the novel's author (Franz Kafka) must be extracted from a specific clause, which can be missed or replaced by a hallucinated answer without context.

How to convert / What changed:
Conversion strategy:
- Add a downstream hop by replacing the original location query with a source-attribution query (ask for the author of the novel the film is based on), while keeping the same anchor entities.

Key modifications:
- S: retain the sentence linking the making-of film to the 1962 film, and retain the clause "based on the novel of the same name by Franz Kafka" as the key evidence.
- Q: rewrite the question to ask for the novel's author rather than the filming location, and describe the film indirectly to require bridging.
- A: ensure the answer is uniquely recoverable from the explicit statement in context.

What to check (evaluation focus):
- The model must correctly resolve the described making-of film to The Trial (1962), then extract the novel author from the "based on the novel ... by Franz Kafka" clause.
- Pitfalls: answering with Orson Welles (director/screenwriter) instead of the novelist; hallucinating another author; failing to bridge from the making-of film to the 1962 film.

\end{Verbatim}
\endgroup
\end{tcolorbox}
\begin{tcolorbox}[
    colback=blue!4,
    colframe=blue!55!black,
    colbacktitle=blue!55!black,
    title=Synthetic Data (Text R\_stress) — Sample 1,
    coltitle=white,
    fonttitle=\bfseries,
    boxrule=1.5pt,
    width=\linewidth,
    before skip=1em,
    after skip=1em,
    breakable,
    left=1mm, right=1mm, boxsep=1mm
]
\begingroup
\footnotesize\ttfamily
\begin{Verbatim}[
  breaklines=true,
  breakanywhere=true,
  breaksymbolleft={},
  breaksymbolright={}
]
==============================
Original Sample (Original QAS)
==============================

S (Context):
Mary Poppins Opens the Door is a British children's fantasy novel by the Australian-British writer P.L. Travers, the third book and last novel in the "Mary Poppins" series that features the magical English nanny Mary Poppins.

Mary Poppins is a fictional character and the eponymous protagonist of P. L. Travers' "Mary Poppins" books and all of their adaptations.

Q (Question):
Which type of character is featured by the P.L. Travers's third book and last novel in the "Mary Poppins" series?

A (Answer):
fictional character


==========================
New Sample (Synthetic QAS)
==========================

S (Context):
"Lanterns of the Silt Sea" is a tidepunk adventure novel by Odran Mavek.
Its protagonist is Vellum Quire, a canal-lantern keeper who can hear drowned songs.

Vellum Quire is a fictional character in the novel "Lanterns of the Silt Sea".
Quire is not to be confused with Vellum Quire, the historical lampwright of Marshgate recorded in guild annals.

Q (Question):
What general type of character is Vellum Quire, who is the protagonist of the novel "Lanterns of the Silt Sea"?

A (Answer):
fictional character


================
Difficulty Card
================

Error Type:
Distractor anchoring on salient traits instead of the requested broad category (evidence omission of the category sentence)

Why it is difficult:
- Broad-category vs trait conflict: the question asks for a general type of character, but the context heavily emphasizes vivid traits (e.g., canal-lantern keeper, hearing drowned songs) that sound like plausible answers.
- Category evidence is easy to miss: the correct label (“fictional character”) appears as a short, explicit classification sentence, which models may omit while copying longer descriptive phrases.
- Light bridge / cross-snippet grounding: the question grounds the entity via its role as protagonist of a named novel, while the explicit category label is stated in a separate entry; the model must cross-reference and prefer the category sentence over trait descriptions.
- Namesake distraction: an additional context introduces a historical lampwright with the same name, encouraging anchoring on occupation/history rather than classification.

How to convert / What changed:
Conversion strategy:
- Replace real entities with fictional ones, and craft contexts that contain (i) a clear broad-category label for the answer and (ii) more salient trait descriptions (plus a namesake) that tempt anchoring.

Key modifications:
- S: add multiple trait-rich descriptions of Vellum Quire (job + supernatural ability) to create a strong distractor.
- S: include a “not to be confused with” namesake to reinforce the temptation to answer with an occupation-like label.
- S: keep one short explicit category statement (“Vellum Quire is a fictional character...”) as the gold evidence.
- Q: explicitly asks for the “general type of character” to enforce choosing the broad category, not a trait.

What to check (evaluation focus):
- Must answer with the broad category “fictional character,” not a trait label like “canal-lantern keeper” or a namesake-related label like “lampwright.”
- Failure signature: copying salient trait phrases while omitting the explicit classification sentence.

(Reference fields for the underlying JSON sample)
supporting_facts:
- Lanterns of the Silt Sea (sent 1)  [protagonist grounding]
- Vellum Quire (sent 0)             [category = fictional character]

pattern_application:
- Core skill: broad-category classification under salient trait distractors.
- Evidence structure: the question grounds the entity in a work ("Lanterns of the Silt Sea"), while the explicit category label (“fictional character”) appears in a separate entry ("Vellum Quire").
- Step 1 (grounding): Use "Lanterns of the Silt Sea" (sent 1) to confirm the referenced protagonist is Vellum Quire (context alignment, not an identity discovery step).
- Step 2 (classification): Use "Vellum Quire" (sent 0) to answer the asked broad category: fictional character.
- Distractor mechanism: multiple contexts emphasize vivid traits (canal-lantern keeper, hearing drowned songs) and a real-sounding namesake ("Marshgate Lampwright Guild"), which can tempt answering with a trait-based label like “canal-lantern keeper” instead of a category.
- Key disambiguator: "Vellum Quire" explicitly states the category and warns “not to be confused with” the historical lampwright, preventing trait/namesake anchoring.
- Target shallow mistake: returning a trait label (e.g., “canal-lantern keeper” / “lampwright”) instead of the general type requested.


\end{Verbatim}
\endgroup
\end{tcolorbox}
\begin{tcolorbox}[
    colback=blue!4,
    colframe=blue!55!black,
    colbacktitle=blue!55!black,
    title=Synthetic Data (Text R\_stress) — Sample 2,
    coltitle=white,
    fonttitle=\bfseries,
    boxrule=1.5pt,
    width=\linewidth,
    before skip=1em,
    after skip=1em,
    breakable,
    left=1mm, right=1mm, boxsep=1mm
]
\begingroup
\footnotesize\ttfamily
\begin{Verbatim}[
  breaklines=true,
  breakanywhere=true,
  breaksymbolleft={},
  breaksymbolright={}
]
==============================
Original Sample (Original QAS)
==============================

S (Context):
The World Summit of Nobel Peace Laureates was initiated by Mikhail Gorbachev in the 90s, as a forum in which the Nobel Peace Laureates and the Peace Laureate Organizations could come together to address global issues with a view to encourage and support peace and human well being in the world.

Mikhail Sergeyevich Gorbachev is a former Soviet statesman.

Q (Question):
What type of forum did a former Soviet statesman initiate?

A (Answer):
Organizations could come together to address global issues


==========================
New Sample (Synthetic QAS)
==========================

S (Context):
The Starwake Exchange was initiated by Kel Varrin as a platform where independent captains could share hazard reports to improve corridor safety.

Kel Varrin is a retired Skyfold Chancellor who stepped down after the Ninth Drift Accord.

Q (Question):
What kind of platform did the retired Skyfold Chancellor initiate?

A (Answer):
a platform where independent captains could share hazard reports to improve corridor safety


================
Difficulty Card
================

Error Type:
Multi-value selection (entity name instead of requested attribute/type)

Why it is difficult:
- Name vs type conflict: the key sentence contains both the platform name (“Starwake Exchange”) and the descriptive type/purpose clause (“a platform where independent captains could share hazard reports…”), and the question asks for the type/kind, not the name.
- Salience trap: the named entity appears first and is short/prominent, making it easy for a model to copy it as an answer span.
- Indirect grounding: the question refers to the creator indirectly (“retired Skyfold Chancellor”), requiring linking that role to Kel Varrin before extracting the platform description.

How to convert / What changed:
Conversion strategy:
- Keep the same sentence pattern “Initiative B was initiated by Person A as a platform/forum where Group C can do Action D to achieve Goal E”, then ask for the “kind/type” so the gold is the descriptive clause.

Key modifications:
- S: place a short initiative name immediately before the descriptive clause to create two plausible answer-like spans in one sentence.
- S: include a separate sentence that maps the role clue (“retired Skyfold Chancellor”) to the person who initiated the platform (Kel Varrin).
- Q: explicitly asks “What kind of platform…” to force selecting the descriptive clause rather than the initiative name.
- A: set gold to the platform’s purpose/type clause, not the initiative’s name.

What to check (evaluation focus):
- Must answer with the descriptive purpose/type clause (“a platform where independent captains could share hazard reports to improve corridor safety”), not “Starwake Exchange”.
- Failure signature: returning the initiative name (or other named entities) instead of the requested “kind/type” description.

(Reference fields for the underlying JSON sample)
supporting_facts:
- Kel Varrin (sent 0)          [role grounding: retired Skyfold Chancellor]
- Starwake Exchange (sent 0)   [platform description clause]

pattern_application:
- Core skill: choose the requested attribute/type when the evidence sentence also contains a highly salient entity name.
- Evidence structure: one context grounds the role → person (“retired Skyfold Chancellor” → Kel Varrin), while another context contains the initiative sentence with both the name and the type/purpose clause.
- Step 1 (grounding): Use “Kel Varrin” (sent 0) to resolve who the “retired Skyfold Chancellor” is.
- Step 2 (attribute extraction): Use “Starwake Exchange” (sent 0) and extract the type/purpose clause (“a platform where independent captains could share hazard reports to improve corridor safety”), not the initiative name.
- Distractor mechanism: the initiative name appears before the clause and looks like a clean answer span, tempting name-copying.
- Target shallow mistake: outputting “Starwake Exchange” instead of the descriptive “kind of platform” phrase.


\end{Verbatim}
\endgroup
\end{tcolorbox}

\subsection{KG Reasoning Instances}

\begin{tcolorbox}[
    colback=blue!4,
    colframe=blue!55!black,
    colbacktitle=blue!55!black,
    title=Synthetic Data (KG K\_stress) — Sample 1,
    coltitle=white,
    fonttitle=\bfseries,
    boxrule=1.5pt,
    width=\linewidth,
    before skip=1em,
    after skip=1em,
    breakable,
    left=1mm, right=1mm, boxsep=1mm
]
\begingroup
\footnotesize\ttfamily
\begin{Verbatim}[
  breaklines=true,
  breakanywhere=true,
  breaksymbolleft={},
  breaksymbolright={}
]
========================
Original Sample (Original QAS)
========================

S (Context):
Answer the following questions using the information in the knowledge graph triples.
knowledge graph triples:Jaxon Bieber people.person.place_of_birth Canada |……| Miami topic_server.population_number 390768 | Jaxon Bieber people.person.gender Male | London location.location.people_born_here Ryan Gosling | Canada location.location.contains Terminal Tower |……| Believe Acoustic music.album.release_date 2013-01-29 | Justin Bieber music.composer.compositions Thought Of You | Thought Of You music.composition.composer Justin Bieber | Singer-songwriter media_common.netflix_genre.titles Sheryl Crow: Wildflower |……

Q (Question):
Who is the brother of the composer of \"Thought Of You\"?

A (Answer):
Jaxon Bieber

========================
New Sample (Synthetic QAS)
========================

S (Context):
Answer the following questions using the information in the knowledge graph triples.
knowledge graph triples:Jaxon Bieber people.person.place_of_birth Canada |……| Miami
topic_server.population_number 390768 | Jaxon Bieber people.person.gender Male | London
location.location.people_born_here Ryan Gosling | Canada location.location.contains
Terminal Tower |……| Believe Acoustic music.album.release_date 2013-01-29 | Justin Bieber music.composer.compositions Thought Of You | Thought Of You music.composition.composer Justin Bieber | Singer-songwriter media_common.netflix_genre.titles Sheryl Crow: Wildflower |……

Q (Question):
Who is the sibling of the person who was born in Canada and whose gender is Male?

A (Answer):
Justin Bieber

========================
Difficulty Card
========================

Error Type: 
Missing Inter-Triple Relation

Root Cause: 
absent kinship linkage in graph structure

Error Details: 
- Both Gold and Model reasoning correctly identify Justin Bieber as the composer of \"Thought Of You\" using the triples (Justin Bieber, music.composer.compositions, Thought Of You) and (Thought Of You, music.composition.composer, Justin Bieber). The divergence occurs at the next required step: deriving the brother of Justin Bieber. The KG contains an entity \"Jaxon Bieber\" with attributes (place_of_birth, gender) but provides no explicit kinship predicate connecting Justin Bieber to any sibling (e.g., no (Justin Bieber, people.person.sibling/brother, Jaxon Bieber) or equivalent). 
- Because this inter-triple connection between the composer entity (Justin Bieber) and a brother entity is missing, the model outputs \"Cannot be determined\" while the gold answer names Jaxon Bieber; however, the KG structure does not support linking Jaxon Bieber to Justin Bieber as a brother.

Missing Knowledge: 
- "Justin Bieber people.person.sibling Jaxon Bieber"
- "Jaxon Bieber people.person.sibling Justin Bieber"

\end{Verbatim}
\endgroup
\end{tcolorbox}
\begin{tcolorbox}[
    colback=blue!4,
    colframe=blue!55!black,
    colbacktitle=blue!55!black,
    title=Synthetic Data (KG K\_stress) — Sample 2,
    coltitle=white,
    fonttitle=\bfseries,
    boxrule=1.5pt,
    width=\linewidth,
    before skip=1em,
    after skip=1em,
    breakable,
    left=1mm, right=1mm, boxsep=1mm
]
\begingroup
\footnotesize\ttfamily
\begin{Verbatim}[
  breaklines=true,
  breakanywhere=true,
  breaksymbolleft={},
  breaksymbolright={}
]
========================
Original Sample (Original QAS)
========================

S (Context):
Answer the following question using the information in the triples. 
knowledge graph triples:……| Montagu House, Bloomsbury architecture.structure.opened 1679 | Montagu House, Bloomsbury type.object.type architecture.building|……| University of Oxford location.location.nearby_airports Gatwick Airport | Wadham College, Oxford education.university.number_of_undergraduates..measurement_unit.dated_integer.number 440 |……| Christ Church, Oxford education.university.number_of_undergraduates..measurement_unit.dated_integer.number 425

Q (Question):
Of the univerisities that were attended by Robert Hooke, which one has the largest number of undergraduates?

A (Answer):
University of Oxford

========================
New Sample (Synthetic QAS)
========================

S (Context):
Answer the following question using the information in the triples. 
knowledge graph triples:……| Montagu House, Bloomsbury architecture.structure.opened 1679 | Montagu House, Bloomsbury type.object.type architecture.building|……| University of Oxford location.location.nearby_airports Gatwick Airport | Wadham College, Oxford education.university.number_of_undergraduates..measurement_unit.dated_integer.number 440 |……| Christ Church, Oxford education.university.number_of_undergraduates..measurement_unit.dated_integer.number 425

Q (Question):
Montagu House, Bloomsbury opened in 1679. Identify the parent organization of the college where Robert Hooke studied, and then state the airport near that parent organization.
      
A (Answer):
Gatwick Airport

========================
Difficulty Card
========================

Error Type: 
Missing Attribute Information

Root Cause: 
Absent education-attendance attributes

Error Details: 
- Comparing the Gold and Model reasoning chains shows both correctly observe that the KG contains no triple linking Robert Hooke to any university via an education/attended relation. The question requires (1) at least one (Robert Hooke, people.person.education/attended, <university>) fact and then (2) an undergraduate-count attribute for those universities to select the maximum. While the KG does contain undergraduate-count attributes for Wadham College, Oxford (440) and Christ Church, Oxford (425), those institutions are not connected to Robert Hooke. 
- The crucial missing information is the education/attendance attribute for the entity Robert Hooke (i.e., a people.person.education-style triple identifying the universities he attended). Because this attribute is absent, the model cannot derive the candidate set of universities attended by Hooke, so it cannot select the largest by undergraduates, leading to disagreement with the provided gold answer ('University of Oxford').
      
Missing Knowledge: 
- "Robert Hooke people.person.education..education.education.institution University of Oxford"
- "University of Oxford education.university.number_of_undergraduates..measurement_unit.dated_integer.number <N>"
- "Robert Hooke people.person.education..education.education.institution Wadham College Oxford"
- "Robert Hooke people.person.education..education.education.institution Christ Church Oxford"
- "Wadham College, Oxford education.educational_institution.parent_organization University of Oxford"
- "Christ Church, Oxford education.educational_institution.parent_organization University of Oxford"
\end{Verbatim}
\endgroup
\end{tcolorbox}

\begin{tcolorbox}[
    colback=blue!4,
    colframe=blue!55!black,
    colbacktitle=blue!55!black,
    title=Synthetic Data (KG R\_stress) — Sample 1,
    coltitle=white,
    fonttitle=\bfseries,
    boxrule=1.5pt,
    width=\linewidth,
    before skip=1em,
    after skip=1em,
    breakable,
    left=1mm, right=1mm, boxsep=1mm
]
\begingroup
\footnotesize\ttfamily
\begin{Verbatim}[
  breaklines=true,
  breakanywhere=true,
  breaksymbolleft={},
  breaksymbolright={}
]
==============================
Original Sample (Original QAS)
==============================

S (Context):
Answer the following questions using the information in the knowledge graph triples. knowledge graph triples: North Korea location.country.currency_formerly_used Korean yen | m.0j5vvhx government.government_position_held.office_holder Kim Jong-un |……| Korean yen finance.currency.countries_used Empire of Japan | North Korea government.governmental_iurisdiction.governing_officials m.0j5vvhx |……| North Korean won finance.currency.countries_used North Korea |……

Q (Question):
Who leads the country that circulates the Korean yen?

A (Answer):
Kim Jong-un

========================
New Sample (Synthetic QAS)
========================

S (Context):
Answer the following question using the information in the triples.
knowledge graph triples: Currency_X currency_formerly_used Country_A | Country_A government.governmental_iurisdiction.governing_officials m.0abc123 | m.0abc123 government.government_position_held.office_holder Leader_A | Currency_X countries_used Country_B | Country_B government.governmental_iurisdiction.governing_officials m.0def456 | m.0def456 government.government_position_held.office_holder Leader_B |……| Country_A located_in Continent_X | Currency_X finance.currency.countries_used Country_B |……| Currency_Y finance.currency.countries_used Country_A | Country_A currency_formerly_used Currency_X | Country_B currency_formerly_used Currency_Y | Country_C currency_formerly_used Currency_Z

Q (Question):
Who is the leader of the country that circulates Currency_X?      

A (Answer):
Leader_A

========================
Difficulty Card
========================

Error Type: 
Path confusion

Root Cause: 
spurious shortest-path preference

Error Details: 
- The gold reasoning uses the intended semantic path: (North Korea, location.country.currency_formerly_used, Korean yen) to identify the relevant country as North Korea, then follows (North Korea, government.governmental_iurisdiction.governing_officials, m.0j5vvhx) and (m.0j5vvhx, government.government_position_held.office_holder, Kim Jong-un) to get the leader. The model instead selects an alternative but semantically misaligned currency→country link: (Korean yen, finance.currency.countries_used, Empire of Japan). 
- This creates a shorter/high-salience path from the currency entity to a "country-like" entity, but it does not match the question's intended interpretation in this KG context (the KG explicitly associates Korean yen with North Korea via the country's former currency). After binding the wrong intermediate entity (Empire of Japan), the model cannot complete the leadership hop because no leader triples exist for Empire of Japan, leading to the incorrect "not specified" answer. Thus the failure is choosing a spurious path among multiple available currency-country connections rather than the gold semantic path.


Transfer conditions (how to inherit): 
- Abstract error form: Given Currency_C, the KG contains two plausible country bindings—(Country_A, currency_formerly_used, Currency_C) and (Currency_C, countries_used, Country_B). The question asks for the leader of the country that circulates/used Currency_C, and the correct path should go through Country_A, but the model instead binds to Country_B and fails or answers incorrectly.

- The difficulty of transfer: This pattern transfers when multiple near-equivalent semantic paths connect the queried entity to different candidate answers. The difficulty is that the model may prefer the more salient/shorter or more directly phrased edge (e.g., 'countries_used') even when the question’s context implies the other edge (e.g., a country-centric 'currency_formerly_used' relation).

- Required characteristics to reproduce: (1) At least two distinct paths from the currency to different country entities using different relations; (2) only one of those countries has an available leader/governing_official chain in the KG; (3) the distractor country either lacks leader info or has unrelated leader info; (4) the question wording is generic enough ('circulates/uses') to make both paths seem plausible, encouraging selection of the wrong path.

\end{Verbatim}
\endgroup
\end{tcolorbox}
\begin{tcolorbox}[
    colback=blue!4,
    colframe=blue!55!black,
    colbacktitle=blue!55!black,
    title=Synthetic Data (KG R\_stress) — Sample 2,
    coltitle=white,
    fonttitle=\bfseries,
    boxrule=1.5pt,
    width=\linewidth,
    before skip=1em,
    after skip=1em,
    breakable,
    left=1mm, right=1mm, boxsep=1mm
]
\begingroup
\footnotesize\ttfamily
\begin{Verbatim}[
  breaklines=true,
  breakanywhere=true,
  breaksymbolleft={},
  breaksymbolright={}
]
========================
Original Sample (Original QAS)
========================

S (Context):
Use the given knowledge graph triples to help you answer the following question. 
knowledge graph triples: South Africa location.country.form_of_government Constitutional republic |……| South Africa food.beer_country_region.beers_from_here S.A. Breweries Castle Lager | South Africa location.country.form_of_government Parliamentary republic……

Q (Question):
Which type of political system, is in the country, where S.A. Breweries Castle Lager beer is produced?

A (Answer):
Constitutional republic, Parliamentary republic

========================
New Sample (Synthetic QAS)
========================

S (Context):
You will be given a list of knowledge graph triples. Answer the following question using the information in the triples.
knowledge graph triples:Product_P1 product.name ViroCure VX-9 | Product_P1 product.category antiviral | Product_P1 product.manufacturer Company_C1 | Product_P1 product.approval.first_approval Event_E1 | Event_E1 event.type regulatory_approval | Event_E1 time.event.start_date 2018-04-12 | Event_E1 event.hosted_in Country_X1 |……| Country_X1 healthcare.system.type Universal coverage | Country_X1 healthcare.system.type Social health insurance……

Q (Question):
Using the triples, first identify the country where the regulatory approval event for ViroCure VX-9 was hosted, then list all healthcare system types recorded for that country.

A (Answer):
Universal coverage, Social health insurance

========================
Difficulty Card
========================

Error Type: 
Attribute-Level Partial Aggregation Error

Root Cause: 
incomplete result aggregation over multiple valid bindings

Error Details: 
- The gold reasoning correctly collects all values for the relation (South Africa, location.country.form_of_government, gov_type) after first identifying South Africa as the country linked to “S.A. Breweries Castle Lager” via (South Africa, food.beer_country_region.beers_from_here, S.A. Breweries Castle Lager). In the KG there are two valid bindings for gov_type: (South Africa, location.country.form_of_government, Constitutional republic) and (South Africa, location.country.form_of_government, Parliamentary republic). 
- The model’s own step 2 acknowledges both triples exist, but in step 3 it outputs only one value (“Parliamentary republic”) instead of aggregating both relation values. Thus, the error is not in entity linking or hop composition, but in failing to return the full set of answers required by the question.

Transfer conditions (how to inherit): 
- Abstract error form: Identify Country_X from a fact like (Country_X, produces_or_associated_with, Product_Y), then query a multi-valued property (Country_X, has_attribute, Value_1) and (Country_X, has_attribute, Value_2 ... Value_n). Gold requires returning {Value_1..Value_n}, but the model returns only a subset.

- The difficulty of transfer: Create contexts where the target relation is explicitly multi-valued (multiple triples with the same subject and predicate) so the solver must aggregate all objects; the common failure mode is top-1/top-k bias or early stopping after finding a salient single value.

- Required characteristics to reproduce: (1) A question whose correct answer is a set/list (e.g., “Which types…?”, “What are the forms…?”). (2) At least two KG triples sharing the same (subject=Country_X, predicate=has_attribute) with different objects. (3) A preceding hop that identifies Country_X from another entity (Product_Y) so the model must both traverse and then enumerate all bindings rather than picking one.
\end{Verbatim}
\endgroup
\end{tcolorbox}

\subsection{Table Reasoning Instances}

\begin{tcolorbox}[
    colback=blue!4,
    colframe=blue!55!black,
    colbacktitle=blue!55!black,
    title=Synthetic Data (Table R\_stress) — Sample 1,
    coltitle=white,
    fonttitle=\bfseries,
    boxrule=1.5pt,
    width=\linewidth,
    before skip=1em,
    after skip=1em,
    breakable,
    left=1mm, right=1mm, boxsep=1mm
]
\begingroup
\footnotesize\ttfamily
\begin{Verbatim}[
  breaklines=true,
  breakanywhere=true,
  breaksymbolleft={},
  breaksymbolright={}
]
==============================
Original Sample (Original QAS)
==============================

S (Context):
Table Header:
Opponent | Goals | Wins | Draws | Losses | +- Goals | Matches | Qualified | Eliminated

Table Rows:
Partizani Tiranë | 28-25 | 13 | 7 | 8 | + 3 | 28 | 9 | 7
Dinamo Tiranë | 30-42 | 8 | 6 | 15 | -12 | 29 | 8 | 8
KS Vllaznia | 25-18 | 13 | 4 | 8 | + 7 | 25 | 9 | 6
KS Flamurtari | 28-13 | 12 | 2 | 3 | +15 | 17 | 8 | 3
KS Teuta | 29-17 | 8 | 9 | 4 | +12 | 21 | 6 | 7
KS Elbasani | 40-17 | 11 | 4 | 4 | +23 | 19 | 8 | 2
KS Besa | 27-26 | 11 | 6 | 7 | + 1 | 24 | 9 | 4
KS Skenderbeu Korce | 18- 5 | 9 | 1 | 1 | +13 | 11 | 7 | 1
KS Tomori | 21- 7 | 7 | 3 | 1 | +14 | 11 | 7 | -
KS Lushnja | 41-15 | 13 | 4 | 5 | +26 | 22 | 8 | 3
Luftëtari Gjirokastër | 19- 3 | 5 | 0 | 0 | +16 | 5 | 3 | -
KS Apolonia | 26- 8 | 8 | 3 | 0 | +18 | 11 | 4 | 1
Besëlidhja Lezhë | 14- 2 | 5 | 0 | 1 | +12 | 6 | 3 | -
KS Kastrioti | 30- 4 | 10 | 1 | 0 | +26 | 11 | 5 | -
KF Naftëtari Kuçovë | 5- 0 | 2 | 0 | 0 | + 5 | 2 | 1 | -
KF Laçi | 13- 4 | 5 | 2 | 1 | + 9 | 8 | 2 | 1
KS Shkumbini | 19- 8 | 6 | 0 | 0 | +11 | 6 | 2 | -
KS Bylis | 11- 4 | 2 | 3 | 1 | + 7 | 6 | 2 | -
KS Sopoti Librazhd | 17- 7 | 2 | 3 | 1 | +10 | 6 | 3 | -
KS Albpetrol | 19- 3 | 5 | 0 | 1 | +16 | 6 | 3 | -
KS Burreli | 1- 2 | 1 | 0 | 1 | - 1 | 2 | 0 | 1
KS Pogradeci | 8- 3 | 2 | 1 | 0 | + 5 | 3 | 2 | -
KS Kamza | 9- 2 | 4 | 0 | 0 | + 7 | 4 | 1 | -
KF Erzeni Shijak | 13- 3 | 6 | 0 | 0 | +10 | 6 | 3 | -
KS Shkëndija | 8- 5 | 2 | 2 | 0 | + 3 | 4 | 2 | -
KS Turbina Cërrik | 13- 3 | 3 | 1 | 0 | +10 | 4 | 2 | -
KF Memaliaj | 9- 2 | 2 | 0 | 0 | + 7 | 2 | 1 | -
KS Tërbuni Pukë | 8- 3 | 3 | 1 | 0 | + 5 | 4 | 2 | -
FK Kukesi | 11- 8 | 2 | 1 | 1 | + 3 | 4 | 1 | 1
KS Iliria | 7- 2 | 2 | 0 | 0 | + 5 | 2 | 1 | -
KF Cakrani | 3- 1 | 1 | 0 | 1 | + 2 | 2 | 1 | -
KS Butrinti Sarandë | 6- 4 | 1 | 1 | 0 | + 2 | 2 | 1 | -
KS Ada Velipojë | 5- 2 | 1 | 1 | 0 | + 3 | 2 | 1 | -
KF Skrapari | 5- 0 | 2 | 0 | 0 | + 5 | 2 | 1 | -
Luzi 2008 | 4- 3 | 1 | 0 | 1 | + 1 | 2 | 1 | -
Dinamo Shkodër | 3- 0 | 1 | 0 | 0 | + 3 | 1 | 1 | -
Garnizoni Durrës | 4- 0 | 1 | 0 | 0 | + 4 | 1 | 1 | -
Albanët | 3- 0 | 1 | 0 | 0 | + 3 | 1 | 1 | -
SK Himarë | 10- 1 | 2 | 0 | 0 | + 9 | 2 | 1 | -
40 opponents* | 592-272 | 195 | 62 | 65 | +320 | 322 | 173 | 44

Q (Question):
how many opponents have the same number of draws as ks flamurtari?

A (Answer):
2


==========================
New Sample (Synthetic QAS)
==========================

S (Context):
Table Header:
Opponent | Week | Venue | Draws | Goals For

Table Rows:
Marrowfen Rovers | Wk 1 | Home | 2 | 3
Duskharbor United | Wk 2 | Away | 1 | 1
Thorncoil Athletic | Wk 3 | Home | 2 | 2
Vellstar City | Wk 4 | Away | 0 | 0
Brineholt FC | Wk 5 | Home | 3 | 4
Kestleford Wanderers | Wk 6 | Away | 2 | 1
Ironvale Crew | Wk 7 | Home | 4 | 5
Cindergate SC | Wk 8 | Away | 1 | 2
Oakhaven Knights | Wk 9 | Home | 2 | 0
Glimmerport FC | Wk 10 | Away | 5 | 6

Q (Question):
How many other opponents have the same number of Draws as Thorncoil Athletic?

A (Answer):
3


================
Difficulty Card
================

Error Type:
Off-by-one self-inclusion in filter-then-count over tables (reference row mistakenly counted)

Why it is difficult:
- “Other opponents” requires exclusion: the target entity is present as a table row, but the question asks for other rows matching its Draws value.
- Equality filtering + aggregation: must (1) read target Draws, (2) filter all rows with that Draws, (3) exclude the target row, (4) count the remainder.
- Typical mistake: count all matches including the target, then return k+1 while explicitly saying “including Thorncoil Athletic.”

How to convert / What changed:
Conversion strategy:
- Build a small synthetic table where the target appears exactly once and exactly k other rows share the same Draws value; ask using “other opponents” to force exclusion.

Key modifications:
- S: include multiple rows with Draws = 2, including the target’s own row (Thorncoil Athletic) plus exactly three other opponents.
- Q: explicitly uses “How many other opponents…” to make exclusion semantically mandatory.
- A: set gold to the number of non-target matching rows (3), not the total matches including the target (4).

What to check (evaluation focus):
- Correct output must be 3 (Marrowfen Rovers, Kestleford Wanderers, Oakhaven Knights), excluding Thorncoil Athletic itself.
- Failure signature: returning 4 by counting the target row along with the three matching opponents.

\end{Verbatim}
\endgroup
\end{tcolorbox}

\begin{tcolorbox}[
    colback=blue!4,
    colframe=blue!55!black,
    colbacktitle=blue!55!black,
    title=Synthetic Data (Table R\_stress) — Sample 2,
    coltitle=white,
    fonttitle=\bfseries,
    boxrule=1.5pt,
    width=\linewidth,
    before skip=1em,
    after skip=1em,
    breakable,
    left=1mm, right=1mm, boxsep=1mm
]
\begingroup
\footnotesize\ttfamily
\begin{Verbatim}[
  breaklines=true,
  breakanywhere=true,
  breaksymbolleft={},
  breaksymbolright={}
]
==============================
Original Sample (Original QAS)
==============================

S (Context):
Table Header:
Record | Athlete | Nation | Venue | Date | #

Table Rows:
4.05 m (13 ft 31⁄4 in) | Sun Caiyun | China | Nanjing, China | 21 May 1992 | 1
4.08 m (13 ft 41⁄2 in) | Sun Caiyun | China | Taiyuan, China | 18 May 1995 | 2
4.08 m (13 ft 41⁄2 in) | Zhong Guiqing | China | Taiyuan, China | 18 May 1995 | 1
4.10 m (13 ft 51⁄4 in) | Daniela Bártová | Czech Republic | Ljubljana, Slovenia | 21 May 1995 | 1
4.12 m (13 ft 6 in) | Daniela Bártová | Czech Republic | Duisburg, Germany | 18 June 1995 | 2
4.13 m (13 ft 61⁄2 in) | Daniela Bártová | Czech Republic | Wesel, Germany | 24 June 1995 | 3
4.14 m (13 ft 63⁄4 in) | Daniela Bártová | Czech Republic | Gateshead, England | 2 July 1995 | 4
4.15 m (13 ft 71⁄4 in) | Daniela Bártová | Czech Republic | Ostrava, Czech Republic | 6 July 1995 | 5
4.16 m (13 ft 73⁄4 in) | Daniela Bártová | Czech Republic | Feldkirch, Austria | 14 July 1995 | 6
4.17 m (13 ft 8 in) | Daniela Bártová | Czech Republic | Feldkirch, Austria | 15 July 1995 | 7
4.18 m (13 ft 81⁄2 in) | Andrea Müller | Germany | Zittau, Germany | 5 August 1995 | 1
4.20 m (13 ft 91⁄4 in) | Daniela Bártová | Czech Republic | Köln, Germany | 18 August 1995 | 8
4.21 m (13 ft 91⁄2 in) | Daniela Bártová | Czech Republic | Linz, Austria | 22 August 1995 | 9
4.22 m (13 ft 10 in) | Daniela Bártová | Czech Republic | Salgotarjan, Hungary | 11 September 1995 | 10
4.25 m (13 ft 111⁄4 in) | Emma George | Australia | Melbourne, Australia | 30 November 1995 | 1
4.28 m (14 ft 01⁄2 in) | Emma George | Australia | Perth, Australia | 17 December 1995 | 2
4.41 m (14 ft 51⁄2 in) | Emma George | Australia | Perth, Australia | 28 January 1996 | 3
4.42 m (14 ft 6 in) | Emma George | Australia | Reims, France | 29 June 1996 | 4
4.45 m (14 ft 7 in) | Emma George | Australia | Sapporo, Japan | 14 July 1996 | 5
4.50 m (14 ft 9 in) | Emma George | Australia | Melbourne, Australia | 8 February 1997 | 6
4.55 m (14 ft 11 in) | Emma George | Australia | Melbourne, Australia | 20 February 1997 | 7
4.57 m (14 ft 113⁄4 in) | Emma George | Australia | Auckland, New Zealand | 21 February 1998 | 8
4.58 m (15 ft 01⁄4 in) | Emma George | Australia | Melbourne, Australia | 14 March 1998 | 9
4.59 m (15 ft 01⁄2 in) | Emma George | Australia | Brisbane, Australia | 21 March 1998 | 10
4.60 m (15 ft 1 in) | Emma George | Australia | Sydney, Australia | 20 February 1999 | 11
4.60 m (15 ft 1 in) | Stacy Dragila | United States | Sevilla, Spain | 21 August 1999 | 1
i 4.60 m (15 ft 1 in) | Stacy Dragila | United States | Pocatello, U.S. | 19 Feb 2000 | 2
i 4.62 m (15 ft 13⁄4 in) | Stacy Dragila | United States | Atlanta, U.S. | 3 Mar 2000 | 3
4.63 m (15 ft 21⁄4 in) | Stacy Dragila | United States | Sacramento, U.S. | 23 July 2000 | 4
i 4.63 m (15 ft 21⁄4 in) | Stacy Dragila | United States | New York City, U.S. | 2 Feb 2001 | 5
i 4.64 m (15 ft 21⁄2 in) | Svetlana Feofanova | Russia | Dortmund, Germany | 11 February 2001 | 1
i 4.66 m (15 ft 31⁄4 in) | Stacy Dragila | United States | Pocatello, U.S. | 17 Feb 2001 | 6
i 4.70 m (15 ft 5 in) | Stacy Dragila | United States | Pocatello, U.S. | 17 Feb 2001 | 7
4.70 m (15 ft 5 in) | Stacy Dragila | United States | Pocatello, U.S. | 27 April 2001 | 8
4.71 m (15 ft 51⁄4 in) | Stacy Dragila | United States | Palo Alto, U.S. | 9 June 2001 | 9
4.81 m (15 ft 91⁄4 in) | Stacy Dragila | United States | Palo Alto, U.S. | 9 June 2001 | 10
4.82 m (15 ft 93⁄4 in) | Yelena Isinbayeva | Russia | Gateshead, England | 13 July 2003 | 1
i 4.83 m (15 ft 10 in) | Yelena Isinbayeva | Russia | Donets'k, Ukraine | 15 February 2004 | 2
i 4.85 m (15 ft 103⁄4 in) | Svetlana Feofanova | Russia | Athens, Greece | 22 February 2004 | 2
i 4.86 m (15 ft 111⁄4 in) | Yelena Isinbayeva | Russia | Budapest, Hungary | 6 March 2004 | 3
4.87 m (15 ft 111⁄2 in) | Yelena Isinbayeva | Russia | Gateshead, England | 27 June 2004 | 4
4.88 m (16 ft 0 in) | Svetlana Feofanova | Russia | Heraklion, Greece | 4 July 2004 | 3
4.89 m (16 ft 01⁄2 in) | Yelena Isinbayeva | Russia | Birmingham, England | 25 July 2004 | 5
4.90 m (16 ft 03⁄4 in) | Yelena Isinbayeva | Russia | London, England | 30 July 2004 | 6
4.91 m (16 ft 11⁄4 in) | Yelena Isinbayeva | Russia | Athens, Greece | 24 August 2004 | 7
4.92 m (16 ft 11⁄2 in) | Yelena Isinbayeva | Russia | Brussels, Belgium | 3 September 2004 | 8
4.93 m (16 ft 2 in) | Yelena Isinbayeva | Russia | Lausanne, Switzerland | 5 July 2005 | 9
4.95 m (16 ft 23⁄4 in) | Yelena Isinbayeva | Russia | Madrid, Spain | 16 July 2005 | 10
4.96 m (16 ft 31⁄4 in) | Yelena Isinbayeva | Russia | London, England | 22 July 2005 | 11
5.00 m (16 ft 43⁄4 in) | Yelena Isinbayeva | Russia | London, England | 22 July 2005 | 12
5.01 m (16 ft 5 in) | Yelena Isinbayeva | Russia | Helsinki, Finland | 9 August 2005 | 13
5.03 m (16 ft 6 in) | Yelena Isinbayeva | Russia | Rome, Italy | 11 July 2008 | 14
5.04 m (16 ft 61⁄4 in) | Yelena Isinbayeva | Russia | Fontvieille, Monaco | 29 July 2008 | 15
5.05 m (16 ft 63⁄4 in) | Yelena Isinbayeva | Russia | Beijing, China | 18 August 2008 | 16
5.06 m (16 ft 7 in) | Yelena Isinbayeva | Russia | Zürich, Switzerland | 28 August 2009 | 17

Q (Question):
which country is the only one to have only one record holder for pole vaulting?

A (Answer):
Germany


==========================
New Sample (Synthetic QAS)
==========================

S (Context):
Table Header:
Meet | Discipline | Guild | Record Holder | Mark

Table Rows:
Frostlamp Trials | Pulse Sprint | Nerith Guild | Velo Quarn | 9.81
Frostlamp Trials | Pulse Sprint | Nerith Guild | Velo Quarn | 9.79
Sablegate Open | Aether Vault | Nerith Guild | Velo Quarn | 18.40
Sablegate Open | Aether Vault | Nerith Guild | Velo Quarn | 18.35
Cindervale Cup | Stone Throw | Brumel Guild | Kara Synn | 71.2
Cindervale Cup | Stone Throw | Brumel Guild | Jori Kest | 72.1
Moonridge Gala | Glass Hurdles | Brumel Guild | Kara Synn | 12.6
Moonridge Gala | Glass Hurdles | Brumel Guild | Jori Kest | 12.4
Rimeharbor Classic | Pulse Sprint | Orrin Guild | Mina Vark | 9.92
Rimeharbor Classic | Pulse Sprint | Orrin Guild | Talo Mern | 9.90

Q (Question):
Which guild is the only one that has only one distinct record holder across all its rows in the table?

A (Answer):
Nerith Guild


================
Difficulty Card
================

Error Type:
Aggregation failure leading to overstrict "unknown" (group-by + COUNT_DISTINCT)

Why it is difficult:
- Unstated-in-row answer: requires grouping by Nation/Guild and counting DISTINCT holders; no single row directly says "only one holder".
- "Only one" uniqueness check: must verify exactly one group has COUNT_DISTINCT(holder)=1, while other groups have COUNT_DISTINCT(holder)>1.
- Failure signature: model outputs "unknown"/refuses despite complete evidence being in-table, often skipping the group-by distinct-count step.
- Repetition trap: many rows for the same group can mislead the model into counting rows instead of distinct holders.

How to convert / What changed:
Conversion strategy:
- Construct a table with multiple rows per group; ensure exactly one group repeats the same holder across all its rows (COUNT_DISTINCT=1), and ensure at least two other groups have >=2 distinct holders.

Key modifications:
- S: Nerith Guild appears multiple times but always with Velo Quarn -> COUNT_DISTINCT(holder)=1.
- S: Brumel Guild has Kara Synn and Jori Kest; Orrin Guild has Mina Vark and Talo Mern -> both have COUNT_DISTINCT>1.
- Q: explicitly asks for the only group with only one distinct record holder.
- A: set gold to the unique qualifying group (Nerith Guild).

What to check (evaluation focus):
- Must compute COUNT_DISTINCT(Record Holder) per group and return the unique group with count=1.
- Typical failure modes: return "unknown"; count total rows not distinct holders; miss that other groups have two different holder names.
\end{Verbatim}
\endgroup
\end{tcolorbox}

\section{Distribution of Root Causes} 
\label{sec:distribution}
\begin{figure*}[t] 
\centering 
\includegraphics[width=\textwidth]{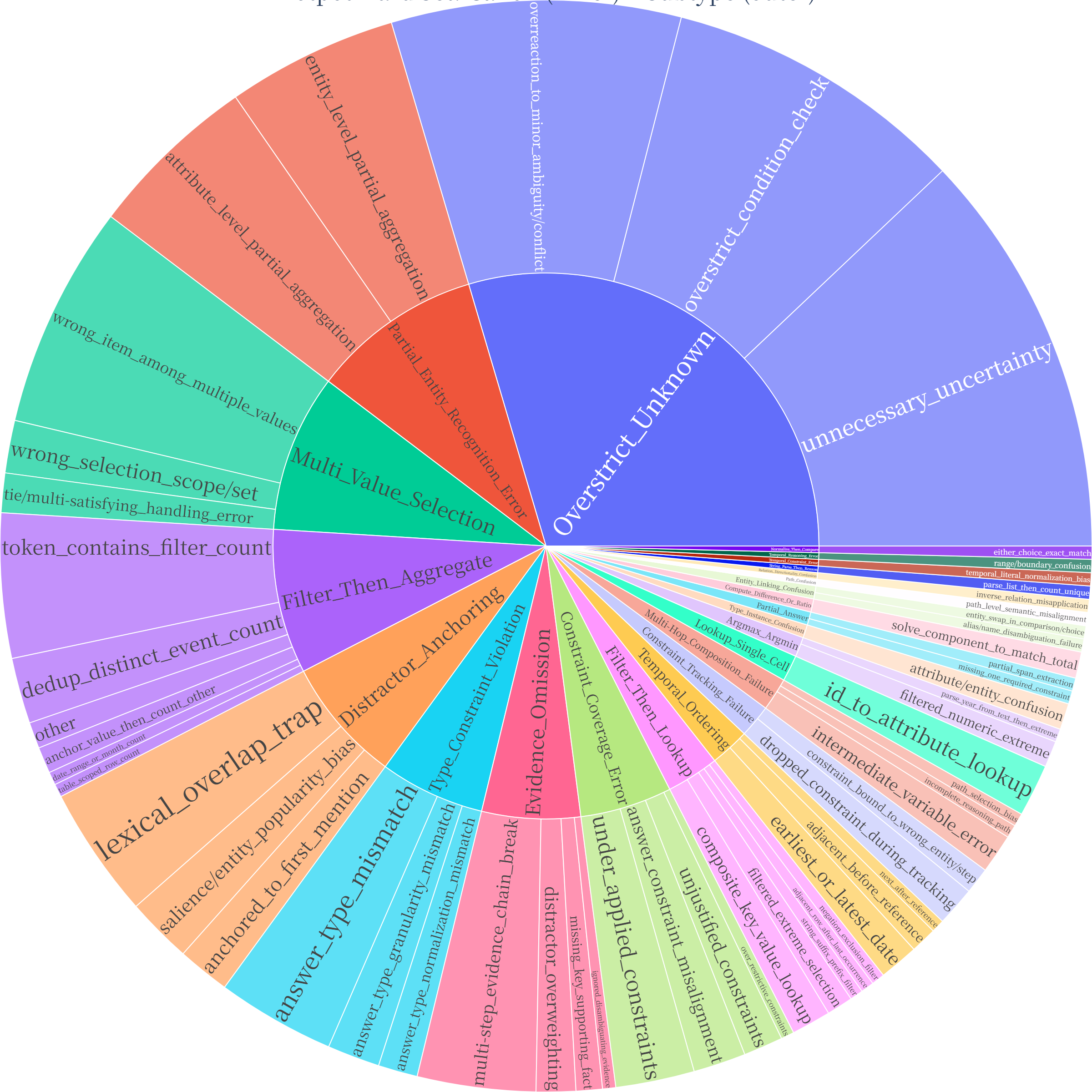} 
\caption{Root-cause distribution among the seed failure cases $\mathcal{E}$.} \label{fig:fig3_enlarged} 
\end{figure*}

\begin{figure*}[t] 
\centering 
\includegraphics[width=\textwidth]{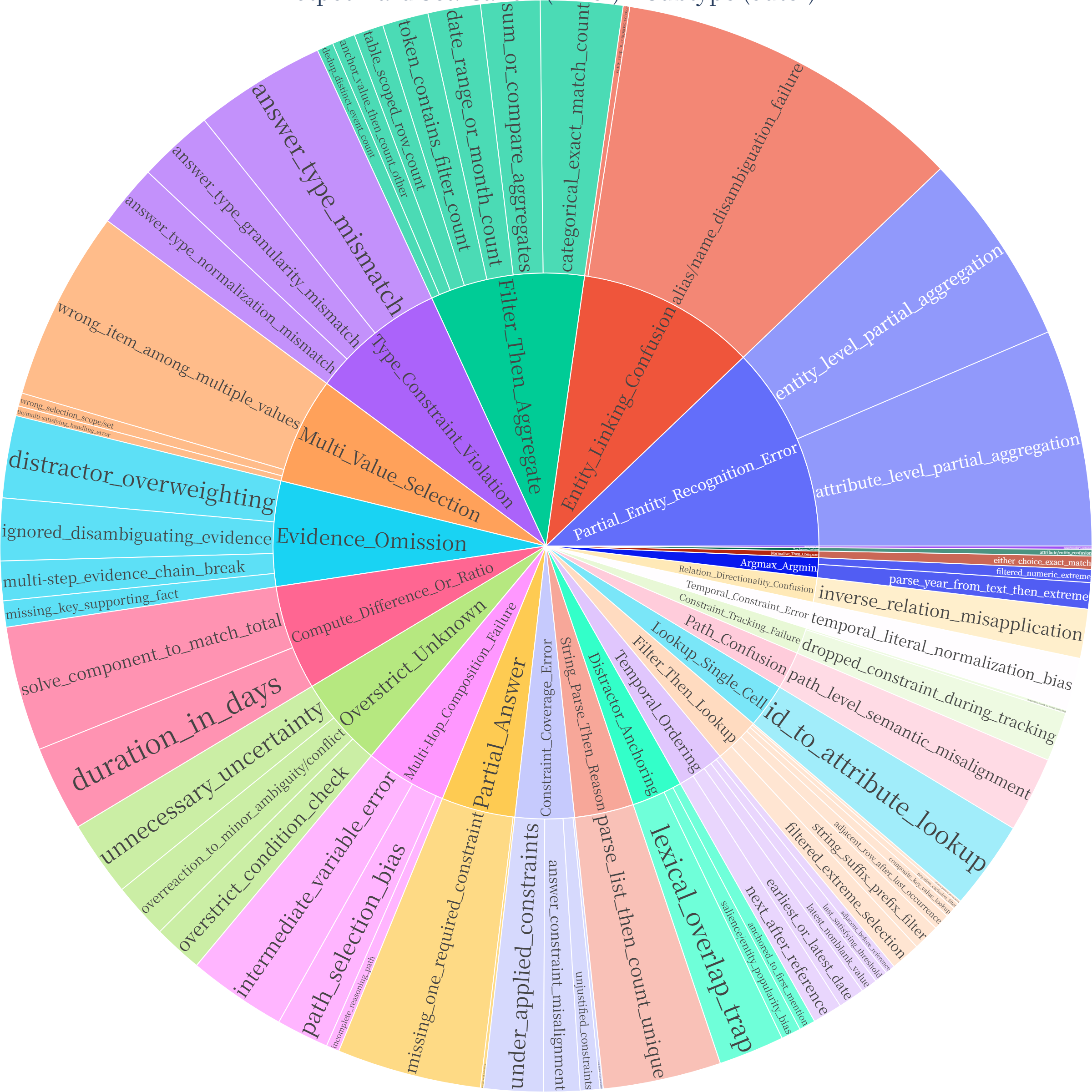} 
\caption{Root-cause distribution of our \textsc{Dynamic-OneEval}, which is comparatively more balanced.} \label{fig:fig3_enlarged} 
\end{figure*}

\end{document}